\documentclass{article}

\usepackage{iclr2027_conference,times}

\usepackage{amsmath,amssymb,amsfonts}
\usepackage{algorithm}
\usepackage{algpseudocode}
\usepackage{graphicx}
\usepackage{textcomp}
\usepackage{xcolor}
\usepackage{booktabs}
\usepackage{multirow}
\usepackage{subcaption}
\usepackage{url}
\usepackage{hyperref}
\usepackage{cleveref}
\hypersetup{
    hidelinks,
    pdftitle={A Hybrid Attention Model Learning Unified Time-aware Patch Representation for Irregular Multivariate Time Series Forecasting},
    pdfauthor={Zhihao Lin; Li Lin; Qi Zhang; Kaiwen Xia; Shuai Wang; Jialin Qiao},
    pdfsubject={Research preprint},
    pdfkeywords={irregular multivariate time series, forecasting, foundation models}
}

\newcommand{\modelname}{\textsc{UniTIMS}}
\newcommand{\modelof}[2][UniTIMS]{#1-#2}
\newcommand{\IMTS}{IMTS}
\newcommand{\projectrepository}{https://github.com/veralily/UniTIMS}
\newcommand{\bestresult}[1]{\textcolor{red}{\textbf{#1}}}
\newcommand{\secondresult}[1]{\textcolor{blue}{\underline{#1}}}

\title{\fontsize{16}{19}\selectfont A Hybrid Attention Model Learning Unified\\
Time-aware Patch Representation for\\
Irregular Multivariate Time Series Forecasting}

\author{
\textbf{Zhihao Lin\textsuperscript{1}, Li Lin\textsuperscript{1}, Qi Zhang\textsuperscript{1}},\\
\textbf{Kaiwen Xia\textsuperscript{2}, Shuai Wang\textsuperscript{1}, Jialin Qiao\textsuperscript{3}}\\[3pt]
\normalfont\textsuperscript{1}Southeast University, China\\
\normalfont\textsuperscript{2}Nanyang Technological University, Singapore\\
\normalfont\textsuperscript{3}Timecho Ltd., China\\[3pt]
\normalfont\texttt{\{zhihao\_lin, linli321, qizhang17, shuaiwang\}@seu.edu.cn}\\
\normalfont\texttt{kaiwen.xia@ntu.edu.sg, jialin.qiao@timecho.com}
}

\iclrfinalcopy

\begin{document}

\maketitle
\pagestyle{plain}
\thispagestyle{plain}

\begin{abstract}
Time series foundation models (TSFMs) have recently delivered impressive zero-shot performance across diverse forecasting tasks. However, real-world decision-making frequently relies on \emph{irregular multivariate time series} (IMTS), where inconsistent inter-observation intervals and asynchronous sampling across variables coexist with informative missingness. 
In this paper, we propose a hybrid attention model that learns a unified time-aware patch representation for IMTS forecasting. We first design a \emph{time-aware patch encoder} that maps a variable number of intra-patch timestamps into a fixed-size embedding, producing a uniform format for irregular patches without resorting to imputation. We then introduce a \emph{time bias attention} mechanism that calibrates inter-patch temporal misalignment and asynchronous cross-channel dependencies as an auxiliary attention offset. Finally, on top of a decoder-only Transformer backbone, we adopt a \emph{hybrid causal mask} that preserves a bidirectional full view over the historical context while keeping the forecast horizon strictly autoregressive. To support large-scale pretraining under irregular settings, we also curate VersaTSA, an archive of $30$B observations that retains the native sampling sparsity of its sources. Experiments on three IMTS benchmarks and a standard regular-MTS benchmark show that our model achieves state-of-the-art zero-shot performance on IMTS and remains competitive when transferred to regular forecasting.
\end{abstract}

\section{Introduction}
With the advancement of foundational models (FMs)~\citep{bommasaniOpportunitiesRisksFoundation2022}, time series analysis is undergoing a profound transformation. Numerous efforts have been made to establish a unified paradigm for Time Series Foundation models (TSFMs) that possess exceptional generalization capabilities for multivariate time series forecasting, such as Sundial~\citep{liuSundialFamilyHighly2025}, Chronos~\citep{ansariChronosLearningLanguage2024}, Moirai~\citep{wooUnifiedTrainingUniversal2024}, and TimesFM~\citep{dasDecoderonlyFoundationModel2024}. However, existing large-scale TSFMs inherently struggle to handle irregularly sampled time series. They require data preprocessing methods such as interpolation to align the data to regular sampling intervals, thereby introducing additional noise~\citep{cheRecurrentNeuralNetworks2018} and obscuring informative missingness~\citep{rubinInferenceMissingData1976}.
This approach leads to mismatches when applying these models to real-world scenarios characterized by inherent temporal inconsistencies. Thus, in this paper, we aim to explore how to build a universal model for irregular multivariate time series (IMTS).
By investigating irregular time series in various real-world domains, such as financial analysis~\citep{sezerFinancialTimeSeries2020}, cyber security~\citep{sousaImprovingIrregularlySampled2020,fawazDeepLearningTime2019}, and weather science~\citep{ravuriSkillfulPrecipitationNowcasting2021}, we can conclude that irregular multivariate time series exhibit two characteristics: (1) \textit{inconsistent time intervals} between observation points and (2) \textit{diverse sampling frequencies} among variables. 
Existing TSFMs rely on traditional positional encoding (e.g., Sinusoidal~\citep{vaswaniAttentionAllYou2023}, RoPE~\citep{suRoFormerEnhancedTransformer2024}, T5 Bias~\citep{press2021train}), which can only encode the sequential order of indices. When temporal intervals are irregular, the positional encoding fails to describe the frequency changes and the natural differences among patches.
Concurrently, some models for irregularly sampled time series address the issue by point-by-point processing~\citep{rubanovaLatentODEsIrregularlySampled2019,luoHiPatchHierarchicalPatch2025, zhangIrregularMultivariateTime2024}. While this approach offers flexibility in handling irregular data, it reduces the information density and increases computational complexity. %

Based on the above observations, our idea starts from a natural understanding of irregular intervals. Empirically, continuous-time encoding transforms timestamps into a continuous space~\citep{kazemiTime2VecLearningVector2019}, enabling models to “perceive” temporal dimensions. Thus, we can build a universal temporal representation learning method based on TSFMs to achieve continuous modeling of time.
To achieve this target, the following challenges need to be addressed.
First, existing continuous-time encoding approaches predominantly embed time using timestamps as units. But multiple timestamps may exist within a single patch. Although the average time encoding provides a solution to represent the temporal information of patches, the capture of \textbf{intricate inner-patch temporal structure} remains underexplored.
Secondly, regarding inter-patch position encoding, conventional methods encode the sequence order of the index. It neglects the \textbf{unevenness between patches}.
Third, existing base models often employ Decoder-Only Transformers with causal masks to maintain the autoregressive manner.
However, inconsistent sampling rates or missing values across different variables lead to overlapping patches during the chunking process, rendering \textbf{causal masking ineffective}. 

To address the above challenges, we propose a hybrid attention model to learn the \textbf{UNI}fied \textbf{T}ime-aware patch embeddings for \textbf{I}rregular \textbf{M}ultivariate Time \textbf{S}eries forecasting ({\modelname}). We specifically decouple the input time series into two parts: the time series values and times. For time series values, we transform them into normalized patch tokens as most TSFMs do. For times, we design a patch-based temporal structure encoding approach, named \textit{time-aware patch encoder}, to represent temporal structural information. It can represent inner-patch temporal information in a universal format. Then, we introduce \textit{time bias attention} to calibrate attention biases arising from inter-patch temporal structural discrepancies. Finally, we introduce the \textit{hybrid causal mask strategy}, full attention for the historical window (input) and causal attention for the autoregressively generated target (future window) based on the common decoder-only architecture used in foundation models. This preserves causal information extraction from irregular time series while preventing unintended exposure of irregular temporal inputs. 

In this work, we conduct a systematic investigation of the pretraining behavior of time series foundation models on irregular multivariate time series (IMTS). By collecting and curating existing time series datasets, we construct a unified dataset collection termed the Versatile Time Series Archive (VersaTSA), which includes 30 billion observations from diverse domains and preserves the irregularity of the originally collected time series. To accommodate pretraining under irregular settings, we reconstruct the timestamps of each sequence and represent time series using sequences of triplets that store the observed time series values, timestamps, and corresponding channels. We summarize our contributions as:
\begin{itemize}
\item We propose a time-aware patch encoding method to represent the irregular temporal information for IMTS, allowing TSFMs to be capable of forecasting irregular time series data without decreasing information density.
\item We present {\modelname}, which pioneers the use of large-scale decoder-only foundation models based on hybrid attention mechanisms to address irregular sequential data with a unified learning of time series values and continuous times.
\item We extend and release VersaTSA, a large-scale time series dataset comprising 30B observations, and use it for pre-training {\modelname}. Experimental results on various datasets show that {\modelname} achieves state-of-the-art performance, outperforming both full-shot time series models and zero-shot results from other TSFMs.
\end{itemize}

\section{Related Work}

Irregular multivariate time series (IMTS) involve asynchronous observations and non-uniform temporal intervals. Early studies mainly treated irregular observations as missing values and relied on kernel methods~\citep{rehfeldComparisonCorrelationAnalysis2011}, multiple imputation~\citep{whiteMultipleImputationUsing2011}, and EM algorithms~\citep{garcia-laencinaPatternClassificationMissing2010}. However, separating imputation from forecasting may overlook informative missing patterns~\citep{wellsStrategiesHandlingMissing2013}. Recent end-to-end methods directly model temporal irregularity, including GRU-D~\citep{cheRecurrentNeuralNetworks2018}, Neural ODE-based approaches~\citep{chenNeuralOrdinaryDifferential2018, rubanovaLatentODEsIrregularlySampled2019,11113235}, and attention-based methods~\citep{shuklaMultiTimeAttentionNetworks2021, zhangIrregularMultivariateTime2024, luoHiPatchHierarchicalPatch2025,11112895} that exploit explicit temporal representations such as Time2Vec~\citep{kazemiTime2VecLearningVector2019}. Despite their effectiveness, these models are generally designed for specific tasks or data distributions, limiting their OOD generalization.

Recent Time Series Foundation Models (TSFMs) have achieved strong zero-shot and few-shot forecasting through large-scale pre-training. Patch-based architectures~\citep{nieTimeSeriesWorth2023} have become prevalent, with models such as Timer~\citep{liuTimerGenerativePretrained2024}, TimesFM~\citep{dasDecoderonlyFoundationModel2024}, and AutoHFormer~\citep{zhang2025autohformerefficienthierarchicalautoregressive} adopting channel-independent modeling, while Chronos-2~\citep{ansariChronos2UnivariateUniversal2025} further captures cross-variate context through synchronous Group Attention. Other TSFMs, including Moirai~\citep{wooUnifiedTrainingUniversal2024}, TimeGPT~\citep{garzaTimeGPT12024}, and Sundial~\citep{liuSundialFamilyHighly2025}, demonstrate broad cross-domain generalization. However, these models are predominantly developed under regular sampling grids: patch indices implicitly correspond to fixed temporal intervals, and synchronous cross-variate modeling assumes aligned observations. Such assumptions prevent them from explicitly capturing asynchronous dependencies and irregular temporal intervals in IMTS.
Motivated by this gap, we aim to extend the foundation-model paradigm to irregular domains by natively modeling asynchronous multivariate contexts and irregular temporal intervals without imputation.

\section{Problem Definition}
Consider an irregular multivariate time series dataset $\mathcal{D} = \{ (u_k, t_k, c_k) \}_{k=1}^{N}$, where $N$ denotes the number of observed variates, $u_k \in \mathbb{R}$ is the observed value, $t_k \in \mathbb{R}^+$ is the timestamp, and  $c_k \in \mathcal{C}$ is the channel identifier. 
Based on the channel partition $\mathcal{C} = \mathcal{C}_\mathcal{Y} \cup \mathcal{C}_\mathcal{Z}$, the dataset $\mathcal{D}$ is decomposed into target variables $\mathcal{Y} = \mathcal{D} \big|_{\mathcal{C}_\mathcal{Y}} = \{ (u,t,c) \mid c \in \mathcal{C}_\mathcal{Y} \}$ and the covariate series $\mathcal{Z} = \mathcal{D} \big|_{\mathcal{C}_\mathcal{Z}}$. The goal of the IMTS forecasting task is to predict the values for a specific query set $Q = \{ (t,c) \mid t \in [t_L,t_{L+H}], c \in \mathcal{C}_\mathcal{Y} \}$, where $[t_L, t_{L+H}]$ defines the future forecast horizon. Formally, 
the IMTS forecasting problem can be formulated as follows:
\begin{equation}
    \mathcal{F}(\mathcal{Y}_{t_1:t_L}, \mathcal{Z}_{t_1:t_{L+H}}, Q) \rightarrow \hat{\mathcal{Y}}_{Q},
\end{equation}
where $\mathcal{Y}_{t_1:t_L} = \{ (u_k, t_k, c_k) \mid c_k \in \mathcal{C}_{\mathcal{Y}}, t_k \in [t_1,t_L]\}$ is the lookback window in the history interval $[t_1, t_L]$, and $\hat{\mathcal{Y}}_{Q} = \{ (u_k, t_k, c_k) \mid (t, c) \in Q\}$ is the set of predicted values for the given queries.
\section{Method}
\begin{figure}[t]
    \centering
    \includegraphics[width=\linewidth,trim={90pt 93pt 248pt 124pt},clip]{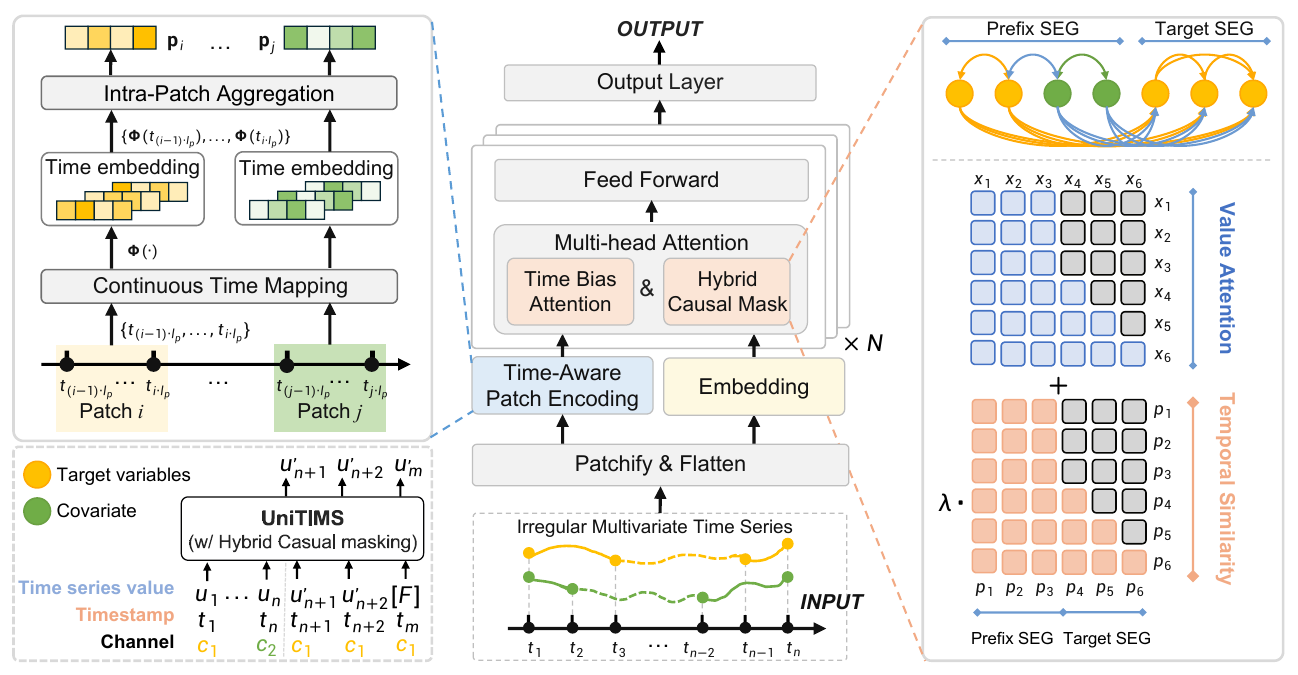}
    \caption{Overview of {\modelname}. The input sequence is split into patches of size 96. Timestamp and value patches are embedded separately and fed into Transformer blocks with Time Bias Attention and a Hybrid Causal Mask, followed by an output layer for value prediction. For simplicity, we illustrate a single target and one observable covariate; multiple variables are flattened and processed identically. During step-by-step inference, predictions at the special token $[\mathtt{F}]$ are fed back, and $[\mathtt{F}]$ advances one step forward.}
    \label{fig:architecture}
\end{figure}

As shown in Figure~\ref{fig:architecture}, {\modelname} models irregular multivariate time series with a patch-based, decoder-only architecture. The input is split into non-overlapping patches~\citep{nieTimeSeriesWorth2023} of length $l_p$. For patch $m$, we form $p_m^{(c)}=\{(u_k,t_k,c)\mid (m-1)l_p+1\le k\le ml_p\}$ and flatten patches into a single sequence~\citep{wooUnifiedTrainingUniversal2024}. For each channel, values $\{u_{(m-1)l_p},\ldots,u_{ml_p}\}$ are encoded into $\mathbf{x}_m$, and timestamps $\{t_{(m-1)l_p},\ldots,t_{ml_p}\}$ into $\mathbf{p}_m$ via time-aware patch encoding (Sec.~\ref{sec:time_aware_patch_encoding}) to capture intra-patch time. These embeddings are fed to a Transformer to autoregressively generate outputs. We replace standard attention with time bias attention (Sec.~\ref{sec:time_bias_attention}) to mitigate inter-patch bias, and use a hybrid causal mask (Sec.~\ref{sec:hybrid_causal_mask}) during generation to model complex dependencies.

\subsection{Time-Aware Patch Encoder}
\label{sec:time_aware_patch_encoding}
Irregular multivariate time series inputs typically consist of two parts: the observed time series values and their corresponding timestamp.
For the time series values, we follow the standard paradigm of mainstream TSFMs, transforming them into continuous patch tokens via patching and normalization. 
Regarding the timestamp, we construct a continuous temporal structure representation that unifies the format for extracting temporal features from irregularly sampled patch-based data.

\subsubsection{Continuous Time Mapping}
We firstly project the original continuous timestamp $t$ into a high-dimensional feature vector to obtain initial time embedding $\mathbf{\Phi}({t}) \in \mathbb{R}^d$: 
\begin{equation}
    \mathbf{\Phi}({t})[i] = \begin{cases} w[i]\cdot t + a[i], & \text{if }i=0 \\a[i] \cdot \sin (w[i] \cdot t), & \text{if }i \text{ is odd}\\a[i] \cdot \cos (w[i] \cdot t), &\text{otherwise}\end{cases},
    \label{eq: mapping}
\end{equation}
where $\mathbf{\Phi}({t})[i]$ is the $i$-th element of $\mathbf{\Phi}({t})$, $d$ denotes the dimension of vector, $w\in \mathbb{R}^d$ and $a \in \mathbb{R}^d$ are learnable frequency and amplitude parameters.
The linear term captures the global temporal evolution trend, while the periodic components are extracted using Fourier terms at different frequencies.
\subsubsection{Intra-Patch Aggregation} To align temporal information with patch representation, given the timestamp set $\mathcal{T}_m = \{t_{k_1}, \cdots, t_{k_{n_m}}\}$ of the $m$-th patch (where $n_m \le l_p$ is the number of \emph{actually observed} timestamps in this patch and may vary across patches and channels under IMTS), we first encode each timestamp via $\mathbf{\Phi}(\cdot)$ to obtain a sequence of $n_m$ time embeddings.
We then aggregate these variable-length embeddings into a single $d$-dimensional patch embedding $\mathbf{p}_m$ by an attention-based pooling that is invariant to the number of valid observations:
\begin{equation}
    \mathbf{p}_m = \sum_{i=1}^{n_m} \alpha_i \, \mathbf{\Phi}(t_{k_i}), \quad
    \alpha_i = \frac{\exp\!\big(\mathbf{q}^\top \mathbf{\Phi}(t_{k_i})\big)}{\sum_{j=1}^{n_m}\exp\!\big(\mathbf{q}^\top \mathbf{\Phi}(t_{k_j})\big)},
    \label{eq:intra_patch_agg}
\end{equation}
where $\mathbf{q}\in\mathbb{R}^d$ is a learnable query vector shared across patches.
For implementation, we materialize each patch as a tensor of length $l_p$ with a binary observation mask $\mathbf{m}_m\in\{0,1\}^{l_p}$ marking the $n_m$ valid positions; unobserved entries are excluded from the softmax in \cref{eq:intra_patch_agg} (mask value $-\infty$ on logits) and contribute neither to the gradient nor to $\mathbf{p}_m$. This design differs fundamentally from imputation- or zero-padding-based encoders: the resulting $\mathbf{p}_m$ depends only on the observed timestamps and their continuous-time embeddings, so patches with different sampling densities map to the \emph{same} representation space without injecting spurious values.
This masked attention pooling effectively handles inconsistencies in sampling frequency and time intervals inherent to irregular multivariate time series, while keeping the output dimension fixed regardless of $n_m$.

\subsection{Time Bias Attention}
\label{sec:time_bias_attention}
In the Transformer backbone, given input value embeddings $\mathbf{x}_i, \mathbf{x}_j \in \mathbb{R}^d$ from patches $i$ and $j$, the initial attention weight matrix $\mathbf{A}_{i,j}^{(l)}$ is calculated as the scaled dot product between the query and key projections, inevitably introducing temporal misalignment bias. 
To address this, we design a time bias attention mechanism that calibrates attention scores by integrating the temporal similarity matrix across patches with both inter-variable and intra-variable correlation matrices. 
Specifically, we first measure the correlation of temporal structures between patches via a bilinear mapping, computing the inter-patch similarity as follows:
\begin{equation}
    \mathbf{G}_{i,j} = \mathbf{p}_i^\top \mathbf{W}_E \mathbf{p}_j,
    \label{eq:time bias attention}
\end{equation}
where $\mathbf{W}_E \in \mathbb{R}^{d \times d}$ is a learnable parameter matrix, and $\mathbf{p}_i$ and $\mathbf{p}_j$ are the time-aware patch embedding corresponding to patches $i$ and $j$. 

Then, to model multivariable correlations, we introduce the any-variate bias~\citep{wooUnifiedTrainingUniversal2024} by constructing the intra-variate matrix $v^{(l)}_1 \mathbb{I}_{\{c_i = c_j\}}$ and the inter-variate matrix $v^{(l)}_2 \mathbb{I}_{\{c_i \neq c_j\}}$, which enhance the attention mechanism’s capacity to represent multivariate time series.
Here, $c_i$ and $c_j$ denote the channel indices of the $i$-th and $j$-th patches, respectively. $\mathbb{I}_{\{\cdot\}}$ is the indicator function used to differentiate intra-channel and inter-channel dependencies. The learnable scalars $v^{(l)}_1$ and $v^{(l)}_2$ allow the model to adaptively balance between channel-independent and channel-mixing modes.

Finally, we incorporate the aforementioned temporal structure prior and multivariate bias as additive inductive biases directly into the initial attention weights. The updated attention matrix is $\hat{\mathbf{A}}_{i,j}^{(l)}$ computed as follows:
\begin{equation}
    \hat{\mathbf{A}}_{i,j}^{(l)} = \mathbf{A}_{i,j}^{(l)} + \lambda^{(l)} \cdot \mathbf{G}_{i,j} + v^{(l)}_1 \mathbb{I}_{\{c_i = c_j\}} + v^{(l)}_2 \mathbb{I}_{\{c_i \neq c_j\}},
    \label{eq:combined attention}
\end{equation}
where $\lambda^{(l)}$ is a learnable scaling factor specific to the $l$-th layer, which dynamically adjusts the influence of the temporal bias $\mathbf{G}$. Overall, we remove positional encodings from the Transformer architecture and instead rely on time bias attention to enable the {\modelname} to perceive inter-patch correlations in irregular multivariate time series modeling.

\subsection{Hybrid Causal Mask}
\label{sec:hybrid_causal_mask}

Based on the updated attention matrix $\hat{\textbf{A}}$, we design a hybrid causal mask (HCM) strategy to maintain autoregressive generation while maximizing the model’s capacity for global modeling of irregular contexts. 
Specifically, given a prefix length $k$, let $\mathbf{x}_i$ and $\mathbf{x}_j$ denote the query and key tokens derived from the value embeddings of their respective patches. The element of the mask matrix $\mathbf{M}_{i, j}$ is defined as: 
\begin{equation}
    \mathbf{M}_{i,j} = \begin{cases} 
    0, & \text{if } j \le \max(i, k) \\ 
    -\infty, & \text{otherwise} 
    \end{cases},
    \label{eq:causal mask}
\end{equation}
Subsequently, incorporating this mask matrix, the final masked attention matrix at the $l$-th layer is computed as:
\begin{equation}
    \mathbf{E}_{i,j}^{(l)} = \mathrm{softmax}(\hat{\mathbf{A}}_{i,j}^{(l)} + \mathbf{M}_{i,j}).
    \label{eq:layer attention}
\end{equation}

The HCM strategy enables the model to perform bidirectional full attention within the prefix segment $(i \le k)$, thereby fully capturing global dependencies in IMTS. In contrast, for the target segment $(i > k)$, the model is strictly constrained by unidirectional causal masking to ensure autoregressive consistency in time series forecasting.
\subsection{VersaTSA Dataset}
\label{sec:versatsa}
\subsubsection{Dataset Description}
To train foundation models tailored for IMTS, we collected and constructed the Versatile Time Series Archive dataset (VersaTSA), designed to support IMTS pretraining by preserving the original sampling sparsity inherent in the data. VersaTSA comprises the following datasets: 
1) Asynchronously sampled torque data from offshore wind power and wave height measurements, with a maximum sampling frequency of 50 Hz, collected over one week.
2) The LOTSA dataset~\citep{wooUnifiedTrainingUniversal2024}, from which padding values have been removed to restore the original irregular sampling distribution.
3) The irregular multivariate time series portion of the Time-IMM dataset~\citep{changTimeIMMDatasetBenchmark2025}, where only the time series data are retained, excluding unstructured textual information.
To facilitate efficient integration with deep learning pipelines, VersaTSA employs Apache Arrow~\citep{arrow} for unified serialized storage. The resulting archive contains over 30 billion observations. Detailed dataset statistics and preprocessing metadata are provided in the project repository~\footnote{\expandafter\url\expandafter{\projectrepository}}.

\subsubsection{Hold-out protocol} To prevent any data leakage in the downstream zero-shot evaluation, the three IMTS evaluation datasets used in Section~\ref{imts_forecasting} (PhysioNet, Human Activity, and USHCN) and the six regular MTS evaluation datasets used in Section~\ref{mts_forecasting} are \emph{strictly excluded} from VersaTSA. We further filter out any sub-corpus from LOTSA whose source overlaps with these benchmarks. Pretraining and downstream evaluation therefore operate on disjoint data sources.

\section{Experiments}

Extensive experiments are conducted to evaluate the performance of {\modelname}, which mainly focus on 1) zero-shot forecasting capability on typical IMTS datasets and general MTS datasets, 2) the ablation study of each module, and 3) the further analysis of attention mechanisms and scalability. Supplementary material, including further discussion of fev-bench results, inference-time comparisons, varying context and horizon settings, training dynamics, and qualitative showcases, is provided in Appendix~\ref{appendix:supplementary_results} and Appendix~\ref{appendix:forecast_visualizations}.

\subsection{Experimental setting}
\label{dataset_and_baseline}
To comprehensively evaluate the performance of {\modelname}, we consider both Irregular Multivariate Time Series (IMTS) forecasting and regular Multivariate Time Series (MTS) forecasting.

\paragraph{IMTS Forecasting} 
We follow prior work~\citep{luoHiPatchHierarchicalPatch2025} and utilize the following three datasets to experiment with models' performance on the IMTS forecasting task: PhysioNet~\citep{silvaPredictingInHospitalMortality2012}, Human Activity~\citep{hadataset}, and USHCN~\citep{menneLongTermDailyMonthly2016}, which span healthcare, biomechanics, and climate science domains. 
We compare {\modelname} against 15 representative baselines, which are divided into three categories: 1) Regular MTS forecasting methods: DLinear~\citep{zengAreTransformersEffective2022}, PatchTST~\citep{nieTimeSeriesWorth2023}, iTransformer~\citep{liuITransformerInvertedTransformers2024}, TimesNet~\citep{wuTimesNetTemporal2DVariation2023}, and TimeMixer~\citep{wangTimeMixerGeneralTime2025}. 2) IMTS forecasting methods: GRU-D~\citep{cheRecurrentNeuralNetworks2018}, CRU~\citep{schirmerModelingIrregularTime2022}, mTAND~\citep{shuklaMultiTimeAttentionNetworks2021}, NeuralFlow~\citep{bilosNeuralFlowsEfficient2021}, Latent-ODE~\citep{rubanovaLatentODEsIrregularlySampled2019}, T-PatchGNN~\citep{zhangIrregularMultivariateTime2024} and Hi-Patch~\citep{luoHiPatchHierarchicalPatch2025}. 3) TSFM methods: Chronos~\citep{ansariChronosLearningLanguage2024}, Chronos-2~\citep{ansariChronos2UnivariateUniversal2025}, Moirai~\citep{wooUnifiedTrainingUniversal2024}, and Sundial~\citep{liuSundialFamilyHighly2025}.

\paragraph{Regular MTS Forecasting} We evaluate the performance of {\modelname} on regular multivariate time series forecasting using the point forecasting benchmark~\citep{liuSundialFamilyHighly2025}, which comprises the 6 datasets, including ETTm1, ETTm2, ETTh1, ETTh2, ECL, Weather, covering 4 different prediction horizons.
We compare {\modelname} against SOTA TSFM baselines: Sundial~\citep{liuSundialFamilyHighly2025}, MOIRAI~\citep{wooUnifiedTrainingUniversal2024}, Chronos~\citep{ansariChronosLearningLanguage2024} and TimesFM~\citep{dasDecoderonlyFoundationModel2024}.

To ensure a thorough and fair comparison, we include all available model sizes for each TSFM baseline that provides multiple variants. For comparability and fairness considerations, three categories of baselines are evaluated under deliberately different protocols. Mean Squared Error (MSE) and Mean Absolute Error (MAE) are used as primary forecasting metrics, computed only over the observed target values and averaged across variables; the full definitions for regular MTS and IMTS are given in Appendix~\ref{appendix:mse_and_mae}.

\begin{table}[t]
  \centering
  \caption{
  Overall performance is evaluated by MSE and MAE (mean $\pm$ std). The best and second-best results are highlighted in \bestresult{red bold} and \secondresult{blue underline}, respectively. Ties are determined at the reported precision. The TSFM model is evaluated under the zero-shot setting. The IMTS forecasting methods use the full-shot setting, and their results are obtained from~\citet{luoHiPatchHierarchicalPatch2025}.}
  \small
  \resizebox{0.95\textwidth}{!}{
    \begin{tabular}{lcccccc}
          \toprule
          ~ & \multicolumn{2}{c}{PhysioNet} & \multicolumn{2}{c}{USHCN} & \multicolumn{2}{c}{Human Activity} \\
          \cmidrule(lr){2-3} \cmidrule(lr){4-5} \cmidrule(lr){6-7}
          ~ & MSE ($\times10^{-3}$) & MAE ($\times10^{-2}$) & MSE ($\times10^{-1}$) & MAE ($\times10^{-1}$) & MSE ($\times10^{-3}$) & MAE ($\times10^{-2}$) \\ 
          \midrule
          {DLinear}      & 41.86 $\pm$ 0.05 & 15.52 $\pm$ 0.03 & 6.21 $\pm$ 0.00 & 3.88 $\pm$ 0.02 & 4.03 $\pm$ 0.01 & 4.21 $\pm$ 0.01 \\ 
          {PatchTST}     & 5.21 $\pm$ 0.33 & 5.10 $\pm$ 0.20 & 5.88 $\pm$ 0.10 & 3.66 $\pm$ 0.13 & 25.56 $\pm$ 4.42 & 10.90 $\pm$ 1.08 \\ 
          {iTransformer} & 3.97 $\pm$ 0.10 & 4.30 $\pm$ 0.08 & 6.17 $\pm$ 0.07 & 4.18 $\pm$ 0.55 & 53.55 $\pm$ 19.59 & 16.87 $\pm$ 4.51 \\ 
          {TimesNet}     & 3.79 $\pm$ 0.05 & 4.28 $\pm$ 0.04 & 5.62 $\pm$ 0.12 & 3.56 $\pm$ 0.12 & 9.30 $\pm$ 0.70 & 5.50 $\pm$ 0.34 \\ 
          {TimeMixer}    & 4.97 $\pm$ 0.31 & 5.02 $\pm$ 0.16 & 5.88 $\pm$ 0.10 & 3.59 $\pm$ 0.07 & 13.98 $\pm$ 0.31 & 6.88 $\pm$ 0.09 \\ 
          \midrule
          {GRU-D}        & 5.76 $\pm$ 0.34 & 4.53 $\pm$ 0.15 & 5.17 $\pm$ 0.06 & 3.21 $\pm$ 0.05 & 3.94 $\pm$ 0.29 & 4.37 $\pm$ 0.21 \\ 
          {CRU}          & 3.03 $\pm$ 0.04 & 3.60 $\pm$ 0.04 & 5.15 $\pm$ 0.50 & 3.18 $\pm$ 0.03 & 6.43 $\pm$ 0.62 & 4.51 $\pm$ 0.16 \\ 
          {mTAND}       & 3.14 $\pm$ 0.09 & 3.71 $\pm$ 0.06 & 5.03 $\pm$ 0.05 & 3.00 $\pm$ 0.06 & 6.18 $\pm$ 0.31 & 4.44 $\pm$ 0.19 \\ 
          {NeuralFlow}   & 4.29 $\pm$ 0.63 & 4.61 $\pm$ 0.43 & 5.41 $\pm$ 0.05 & 3.35 $\pm$ 0.06 & 7.68 $\pm$ 0.37 & 4.84 $\pm$ 0.19 \\ 
          {Latent-ODE}   & 3.32 $\pm$ 0.10 & 3.91 $\pm$ 0.08 & 5.16 $\pm$ 0.04 & 3.21 $\pm$ 0.07 & 6.85 $\pm$ 0.28 & 4.77 $\pm$ 0.17 \\ 
          {T-PatchGNN}   & 2.79 $\pm$ 0.09 & 3.24 $\pm$ 0.06 & 5.00 $\pm$ 0.03 & 3.07 $\pm$ 0.05 & 5.06 $\pm$ 0.10 & 3.75 $\pm$ 0.07 \\ 
          {Hi-Patch}   & 2.57 $\pm$ 0.02 & 3.11 $\pm$ 0.03 & 4.94 $\pm$ 0.05 & 2.96 $\pm$ 0.04 & 4.86 $\pm$ 0.03 & 3.62 $\pm$ 0.07 \\ 
          \midrule
          {Chronos-2}   & 3.94 $\pm$ 0.02 & 4.12 $\pm$ 0.04 & 4.71 $\pm$ 0.06 & 2.83 $\pm$ 0.03 & 4.65 $\pm$ 0.10 & 3.54 $\pm$ 0.07 \\ 
          {\modelof[Moirai]{Base}}       & 4.31 $\pm$ 0.30 & 4.71 $\pm$ 0.43 & 6.04 $\pm$ 0.10 & 4.11 $\pm$ 0.14 & 5.72 $\pm$ 0.06 & 4.63 $\pm$ 0.10 \\ 
          {\modelof[Moirai]{Large}}       & 4.19 $\pm$ 0.26 & 4.27 $\pm$ 0.04 & 6.00 $\pm$ 0.11 & 4.09 $\pm$ 0.12 & 5.68 $\pm$ 0.13 & 5.59 $\pm$ 0.10 \\ 
          {\modelof[Sundial]{Base}}      & 4.79 $\pm$ 0.02 & 4.91 $\pm$ 0.14 & 5.78 $\pm$ 0.06 & 3.83 $\pm$ 0.05 & 5.79 $\pm$ 0.12 & 5.67 $\pm$ 0.33 \\           
          {\modelof[Sundial]{Large}}      & 4.76 $\pm$ 0.04 & 4.82 $\pm$ 0.13 & 5.65 $\pm$ 0.04 & 3.76 $\pm$ 0.04 & 5.76 $\pm$ 0.13 & 5.57 $\pm$ 0.31 \\           
          \midrule
         {\modelof{Small}} & 3.41 $\pm$ 0.04 & 3.89 $\pm$ 0.05 & 4.92 $\pm$ 0.03 & 2.91 $\pm$ 0.04 & 4.63 $\pm$ 0.12 & 3.44 $\pm$ 0.06 \\ 
          {\modelof{Base}}  & \bestresult{2.37 $\pm$ 0.03} & \bestresult{2.74 $\pm$ 0.03} & \secondresult{3.62 $\pm$ 0.03} & \bestresult{2.29 $\pm$ 0.03} & \secondresult{3.41 $\pm$ 0.10} & \secondresult{2.07 $\pm$ 0.07} \\
          {\modelof{Large}} & \secondresult{2.52 $\pm$ 0.03} & \secondresult{3.07 $\pm$ 0.03} & \bestresult{3.53 $\pm$ 0.02} & \secondresult{2.32 $\pm$ 0.02} & \bestresult{2.44 $\pm$ 0.11} & \bestresult{1.61 $\pm$ 0.05} \\
          \bottomrule
        \end{tabular}
    }
  \label{tab:main-result}%
\end{table}%

\begin{table}[ht]
  \centering
  \caption{Point forecasting results of regular MTS. Results are averaged across prediction lengths {96, 192, 336, 720}. Best results are highlighted in \bestresult{red bold}, and second-best results are highlighted in \secondresult{blue underline}. Ties are determined at the reported precision. Baseline results are obtained from~\citet{liuSundialFamilyHighly2025}.}
  \label{tab:mts-result}
  \small
  \resizebox{0.85\textwidth}{!}{
    \begin{tabular}{lcccccccccccc}
      \toprule
      ~ & \multicolumn{2}{c}{ETTm1} & \multicolumn{2}{c}{ETTm2} & \multicolumn{2}{c}{ETTh1} & \multicolumn{2}{c}{ETTh2} & \multicolumn{2}{c}{ECL} & \multicolumn{2}{c}{Weather} \\
      \cmidrule(lr){2-3} \cmidrule(lr){4-5} \cmidrule(lr){6-7} \cmidrule(lr){8-9} \cmidrule(lr){10-11} \cmidrule(lr){12-13}
      ~ & MSE & MAE & MSE & MAE & MSE & MAE & MSE & MAE & MSE & MAE & MSE & MAE \\
      \midrule
      TimesFM & 0.433 & 0.418 & 0.328 & 0.346 & 0.473 & 0.443 & 0.392 & 0.406 & - & - & - & - \\
       \midrule
      \modelof[Chronos]{Base} & 0.645 & 0.500 & 0.310 & 0.350 & 0.591 & 0.468 & 0.405 & 0.410 & 0.214 & 0.278 & 0.292 & 0.315 \\
      \modelof[Chronos]{Large} & 0.555 & 0.465 & 0.295 & 0.338 & 0.588 & 0.466 & 0.455 & 0.427 & 0.204 & 0.273 & 0.279 & 0.306 \\
      \midrule
      \modelof[Moirai]{Base} & 0.406 & 0.385 & 0.311 & 0.337 & 0.417 & \secondresult{0.419} & 0.362 & 0.382 & 0.187 & 0.274 & 0.287 & 0.281 \\
      
      \modelof[Moirai]{Large} & 0.422 & 0.391 & 0.329 & 0.343 & 0.480 & 0.439 & 0.367 & 0.377 & 0.186 & 0.270 & 0.264 & \secondresult{0.273} \\
       \midrule
      \modelof[Sundial]{Base} & \secondresult{0.336} & 0.377 & 0.258 & 0.320 & 0.411 & 0.434 & \bestresult{0.333} & 0.387 & 0.169 & 0.265 & \bestresult{0.234} & \bestresult{0.270} \\
      \modelof[Sundial]{Large} & \bestresult{0.331} & \bestresult{0.369} & \secondresult{0.254} & \bestresult{0.315} & \bestresult{0.395} & 0.420 & \secondresult{0.334} & 0.387 & \secondresult{0.166} & 0.262 & \secondresult{0.238} & 0.275 \\
      \midrule
      \modelof{Small} & 0.400 & 0.381 & 0.283 & 0.331 & 0.431 & 0.436 & 0.354 & 0.378 & 0.169 & 0.269 & 0.261 & 0.283 \\
      \modelof{Base} & 0.384 & 0.374 & 0.271 & \secondresult{0.318} & 0.420 & 0.422 & 0.351 & \bestresult{0.374} & 0.167 & \secondresult{0.260} & 0.254 & 0.279 \\
      \modelof{Large} & 0.362 & \secondresult{0.371} & \bestresult{0.253} & \bestresult{0.315} & \secondresult{0.403} & \bestresult{0.418} & 0.352 & \secondresult{0.376} & \bestresult{0.162} & \bestresult{0.257} & 0.241 & 0.274 \\
      \bottomrule
    \end{tabular}
  }
\end{table}

\subsection{Implement Details}
\label{implement_details}
We implement {\modelname} in three configurations: Small (60M), Base (403M), and Large (1.3B); the detailed layer and dimension settings are reported in Appendix~\ref{appendix:model_configurations}. All models are trained with the AdamW optimizer under a linear-warmup cosine-decay schedule on NVIDIA H100-NVL GPUs, with full optimization, sampling, and inference details provided in Appendix~\ref{appendix:implementation_detail}.

\subsection{Overall Performance}
\subsubsection{{\IMTS} Forecasting}
\label{imts_forecasting}
Table~\ref{tab:main-result} reports the zero-shot forecasting performance on three IMTS datasets. The experimental results show that \modelof{Base} and \modelof{Large} achieve superior performance, consistently ranking best or second-best across all evaluation metrics, with several findings: 1) The \modelof{Small} using fewer parameters achieves the averaged MSE reduction of 3.14\% and the averaged MAE reduction of 1.86\% compared to the state-of-the-art TSFM baseline, Chronos-2. This improvement is mainly attributed to the widespread use of positional encodings in existing TSFMs, which limits their ability to effectively capture the inherent irregularities and inter-variable dependencies in IMTS. 2) The zero-shot forecast results of \modelof{Base} and \modelof{Large} surpass those of specialized full-shot models extensively trained on the target datasets, demonstrating that leveraging high-quality, large-scale data such as VersaTSA can effectively activate model parameters and enhance the generalization performance of TSFMs.

To further evaluate robustness under different observation-to-forecast ratios, we evaluate {\modelname} on Human Activity and PhysioNet under both balanced and short-history, long-horizon configurations, with full results reported in Appendix~\ref{appendix:varying_context_horizon} (Tables~\ref{tab:ha_varying_horizon} and~\ref{tab:physionet_varying_horizon}). The proposed variants maintain strong performance across all settings, indicating stable adaptability to shifting temporal windows.

\subsubsection{Regular MTS Forecasting}
\label{mts_forecasting}
Table~\ref{tab:mts-result} reports regular MTS forecasting performance averaged over the four prediction horizons in the point forecasting benchmark, with the full per-horizon results deferred to Appendix~\ref{appendix:full_results_of_MTS}. Remarkably, {\modelname} achieves state-of-the-art average performance on the ECL dataset and competitive performance on ETTh2. It is important to note that our model was pre-trained on the irregular VersaTSA dataset without the aid of synthetic generation tools like KernelSynth~\citep{ansariChronosLearningLanguage2024}, resulting in a significantly smaller training corpus than state-of-the-art regular TSFMs. 
Meanwhile, \modelof{Large} achieves the lowest reported average MSE on ETTm2 and ties with Sundial-Large for the lowest average MAE at the reported precision.
\par
\begin{figure}[t]
    \centering
    \includegraphics[width=\linewidth]{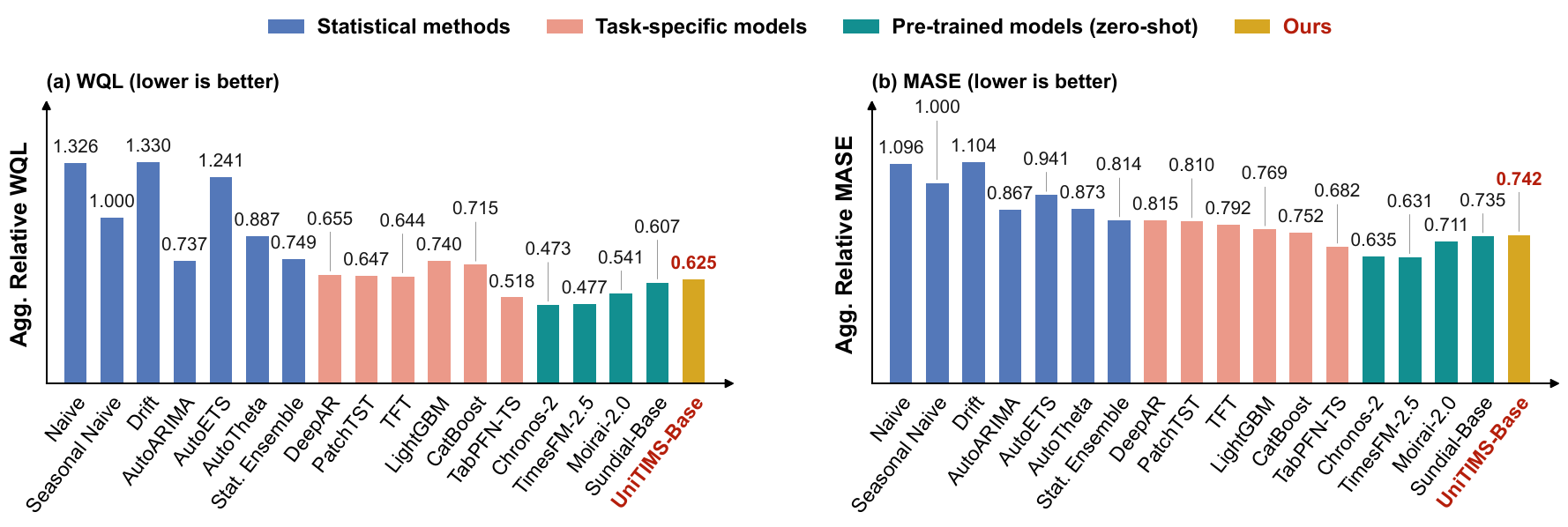}
    \caption{Fev-bench results with (a) aggregate relative WQL and (b) aggregate relative MASE. Colors distinguish statistical methods, task-specific models, and pre-trained models evaluated zero-shot. \modelof{Base} is shown in gold, with its name and scores highlighted in red bold. Lower values indicate better performance.}
    \label{fig:fev-bench}
    \vspace{-10pt}
\end{figure}

On the open leaderboard fev-bench (100 forecasting tasks, none seen during training), {\modelname} demonstrates robust and trustworthy performance despite being designed for irregular time series, and all pre-trained models under zero-shot inference show clear superiority over task-specific and statistical baselines. Figure~\ref{fig:fev-bench} reports the aggregate relative WQL and MASE results, with further discussion in Appendix~\ref{appendix:fev_bench}. We further compare the wall-clock inference time of representative TSFMs under the same hardware configuration in Appendix~\ref{appendix:inference_time}: {\modelname} achieves competitive inference speed relative to foundation models of comparable scale, indicating that the time-aware hybrid attention design does not introduce prohibitive computational overhead.

\subsection{Ablation Study}
\begin{figure}[ht]
    \centering
    \includegraphics[width=0.8\linewidth]{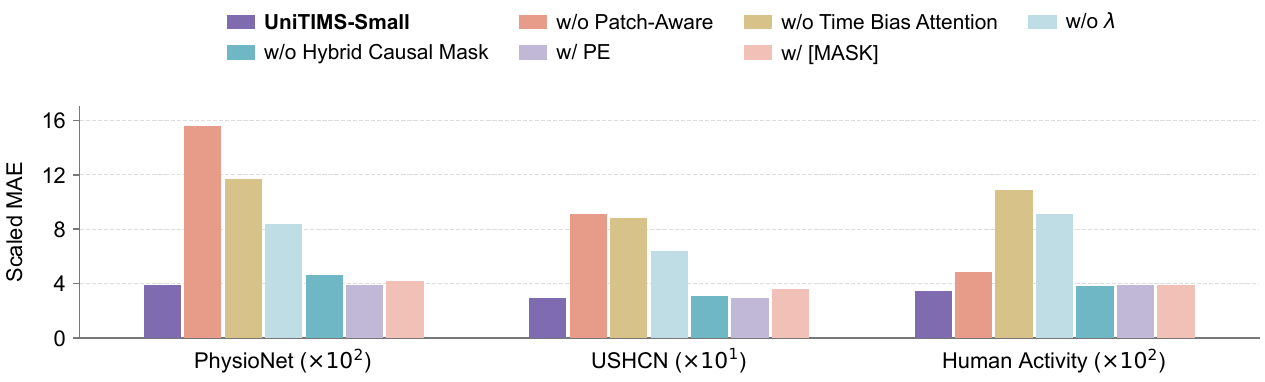}
    \caption{Ablation results of \modelof{Small} on three IMTS datasets. MAE is scaled as indicated on the horizontal axis; lower is better. PE denotes sinusoidal positional encoding; \texttt{[MASK]} is a special token used to supply query information at corresponding positions.}
    \label{fig:ablation_study}
    \vspace{-8pt}
\end{figure}

We present a series of ablation results in Figure~\ref{fig:ablation_study} on PhysioNet, Human Activity, and USHCN datasets, starting from the default \modelof{Small}. Here, \textbf{w/ PE} adds sinusoidal positional encoding (PE), while \textbf{w/ [MASK]} uses a special \texttt{[MASK]} token to supply query information at the corresponding positions. First, we ablate the Patch-Aware Time Embedding component (\textbf{w/o Patch-Aware}). Instead of concatenating the vector representations of multiple timestamps within a patch and projecting them via a linear layer, we employed the average of these vector representations to represent the temporal features of the patch. We observed a significant deterioration in normalized MAE when removing the Patch-aware time embedding. Subsequently, we conducted ablation on the Time Bias Attention module (\textbf{w/o Time Bias Attention}), again noting an increase in normalized MAE, indicating diminished model performance. Next, we removed the learnable layer-specific scaling factor $\lambda$ of the temporal bias within Time Bias Attention (\textbf{w/o $\lambda$}), observing a slight shift in MAE. Finally, we conducted ablation experiments on the Hybrid Causal Mask (\textbf{w/o Hybrid Causal Mask}). Upon replacing it with the Full Attention Mask, experimental results demonstrated a significant decline in the model's prediction accuracy.
From the ablation study, we can conclude that each component in our framework contributes to the final improvements for irregular multivariate time series forecasting tasks.
\label{ablation_study}

We observed that introducing additional positional information via PE did not yield significant performance gains for the model. Furthermore, the \textbf{w/ [MASK]} variant demonstrated that this backfilling approach exhibited a suppressing effect on the model's capabilities.

\subsection{Hybrid Causal Mask Analysis}
\label{sec:further_analysis}
\begin{figure}[h]
\begin{center}
    \centerline{\includegraphics[width=0.8\linewidth]{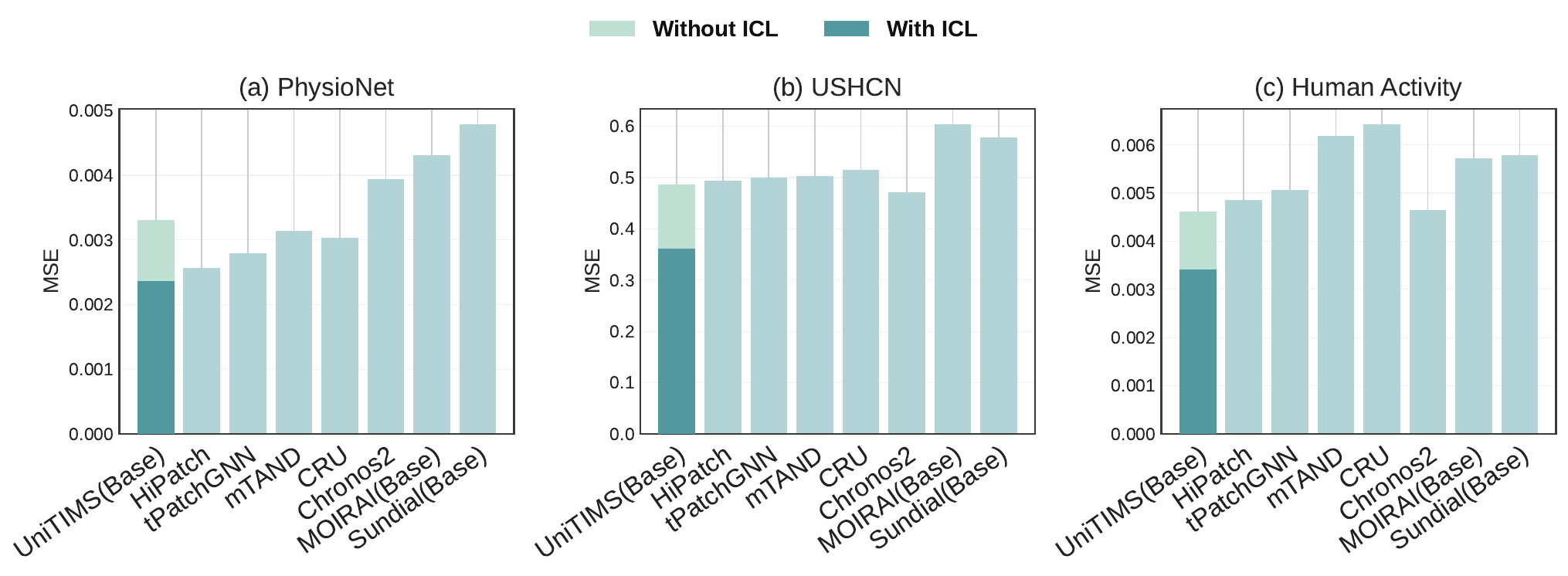}}
    \caption{The analysis of Hybrid Causal Mask on three datasets of IMTS. Evaluated by MSE, where lower values indicate better performance.}
    \label{fig:ICL}
\end{center}
\vspace{-10pt}
\end{figure}

We design a comparative study to examine the contextual modeling capability of the proposed Hybrid Causal Mask.
Specifically, we compare two settings for IMTS forecasting:
(i) \textit{without ICL} by replacing the Hybrid Causal Mask with a standard Causal Attention Mask, where multiple variate time series are provided in a channel-independent (CI) manner; 
and
(ii) \textit{with ICL} by adopting the Hybrid Causal Mask and concatenating multiple multivariate time series into a single sequence as contextual input.
Figure~\ref{fig:ICL} reports the results evaluated by MSE. Across three irregularly sampled time-series datasets, the Hybrid Causal Mask consistently achieves better performance, indicating a stronger ability to leverage contextual information.

Qualitative zero-shot forecasting visualizations of {\modelof{Large}} on both IMTS and regular MTS datasets are provided in Appendix~\ref{appendix:forecast_visualizations}, where the predicted curves preserve the overall evolution of the ground truth under irregular intervals, asynchronous observations, and varying frequencies.

\section{Conclusion}
We introduce {\modelname}, a time-aware decoder-only Transformer with a hybrid causal mask for irregular multivariate time series forecasting (IMTS), with a particular focus on informative missingness and complex contextual dependencies issues in IMTS. We also introduced VersaTSA, a large collection of IMTS data for pretraining. Experiments on several datasets and benchmarks demonstrate {\modelname} achieves superior zero-shot performance on IMTS and remains competitive on regular multivariate time series.
Despite its strong out-of-the-box performance on IMTS, the inference efficiency of {\modelname} remains an
important practical consideration. Incorporating inference
acceleration techniques like KV cache may significantly improve its efficiency and scalability.

\subsection*{Reproducibility statement}
The model architecture and training objective are described in
Sections~\ref{sec:time_aware_patch_encoding}--\ref{sec:hybrid_causal_mask} and
Section~\ref{sec:versatsa}. Dataset and baseline choices, evaluation protocols,
metrics, model configurations, optimization hyperparameters, and inference
details are reported in Section~\ref{dataset_and_baseline} and
Section~\ref{implement_details}. The project repository referenced in
Section~\ref{sec:versatsa} provides additional material for reproducing the
data-processing pipeline.

\subsection*{AI use statement}
During preparation of this manuscript version, generative AI tools assisted with
\LaTeX{} formatting, figure and table presentation, consistency checks, and
copy-editing. The authors are responsible for the scientific content and for
verifying all results and references.

\nocite{zheng2026causalprescalableeffectivedata}

\appendix
\section{Versatile Time Series Archive}
\label{appendix:VersaTSA}
Versa collects multiple datasets of both regular MTS and IMTS, and stores them in the form of three lists $(U,T,C)$ to facilitate pretraining of TSFMs in IMTS. For regular MTS datasets, we filter out missing values and reconstruct the timestamps of each sequence using the provided start field. We further details on key attributes of each dataset, including the domain, the number of time series, and the number of observations. And the frequency is reported for regular MTS.
\begin{table}[ht!]
  \centering
  \caption{Datasets and key properties from LOTSA.}
  \label{tab:LOTSA}
  \begin{tabular}{lcccc}
    \toprule
    \textbf{Dataset} & \textbf{Domain} & \textbf{Frequency} & 
    \textbf{\# Time Series} & 
    \textbf{\# Obs.} \\
    \midrule
    BDG-2 Panther & Energy & H     & 105      & 919,800 \\
    BDG-2 Fox & Energy & H     & 135      & 2,324,568 \\
    BDG-2 Rat & Energy & H     & 280      & 4,728,288 \\
    BDG-2 Bear & Energy & H     & 91       & 1,482,312 \\
    Low Carbon London & Energy & H        & 0     & 9,543,348 \\
    SMART & Energy & H     & 5        & 95,709 \\
    IDEAL & Energy & H     & 219      & 1,265,672 \\
    Sceaux & Energy & H     & 1        & 34,223 \\
    Borealis & Energy & H     & 15       & 83,269 \\
    Buildings900K & Energy & H     & 1,792,328    & 15,702,590,000 \\ 
    \midrule
    Azure VM Traces 2017 & CloudOps & 5T    & 159,472    & 885,522,908 \\
    Borg Cluster Data 2011 & CloudOps & 5T    & 143,386    & 537,552,854 \\
    Alibaba Cluster Trace 2018 & CloudOps & 5T    & 58,409    & 95,192,530 \\
    \midrule
    Taxi  & Transport & 30T   & 67,984    & 54,999,060 \\
    Uber TLC Daily & Transport & D     & 262      & 47,087 \\
    Uber TLC Hourly & Transport & H     & 262      & 1,129,444 \\
    Wiki-Rolling & Web   & D     & 47,675    & 40,619,100 \\
    M5    & Sales & D     & 30,490    & 58,327,370 \\
    \midrule
    LargeST & Transport & 5T    & 42,333    & 4,452,510,528 \\
    \midrule
    PEMS03 & Transport & 5T    & 358      & 9,382,464 \\
    PEMS04 & Transport & 5T    & 307      & 5,216,544 \\
    PEMS07 & Transport & 5T    & 883      & 24,921,792 \\
    PEMS08 & Transport & 5T    & 170      & 3,035,520 \\
    PEMS Bay & Transport & 5T    & 325      & 16,937,700 \\
    Los-Loop & Transport & 5T    & 207     & 7,094,304 \\
    Loop Seattle & Transport & 5T    & 323      & 33,953,760 \\
    SZ-Taxi & Transport & 15T   & 156     & 464,256 \\
    Beijing Subway & Transport & 30T   & 276      & 248,400 \\
    SHMetro & Transport & 15T   & 288      & 1,934,208 \\
    HZMetro & Transport & 15T   & 80        & 146,000 \\
    Rotterdam & Transport & 2T    & 208       & 4,813,536 \\
    Q-Traffic & Transport & 15T   & 45,148     & 264,386,688 \\
    \bottomrule
    \end{tabular}%
  \label{tab:lotsa_1}%
\end{table}
\begin{table}[htp]
  \centering
  \ContinuedFloat
  \caption{Cont. Datasets and key properties from LOTSA.}
  \begin{tabular}{lcccc}
    \toprule
    \multicolumn{5}{c}{... Continued from previous page}\\
    \midrule
    \textbf{Dataset} & \textbf{Domain} & \textbf{Frequency} & 
    \textbf{\# Time Series} & 
    \textbf{\# Obs.} \\
    \midrule
    KDD Cup 2022   & Energy & 10T   & 134      & 4,727,519 \\
    GoDaddy & Econ/Fin & M     & 3,135   & 128,535 \\
    Favorita Sales & Sales & D     & 111,840  & 139,179,538 \\
    Favorita Transactions & Sales & D     & 54     & 84,408 \\
    Restaurant & Sales & D     & 216     & 76,573 \\
    Hierarchical Sales   & Sales & D     & 118      & 212,164 \\
    China Air Quality  & Nature & H     & 437     & 5,739,234 \\
    Beijing Air Quality  & Nature & H     & 12      & 420,768 \\
    Residential Load Power   & Energy & T     & 271    & 145,994,559 \\
    Residential PV Power     & Energy & T     & 233       & 125,338,950 \\
    CDC Fluview ILINet   & Healthcare & W     & 75       & 63,903 \\
    CDC Fluview WHO NREVSS   & Healthcare & W     & 74      & 41,760 \\
    Project Tycho  & Healthcare & W     & 1,258  & 1,377,707 \\
    \midrule
    London Smart Meters & Energy & 30T   & 5,520  & 166,238,880 \\
    Wind Farms & Energy & T     & 337   & 172,165,370 \\
    Wind Power & Energy & 4S    & 1      & 7,397,147 \\
    Solar Power & Energy & 4S    & 1     & 7,397,222 \\
    Oikolab Weather & Climate & H     & 8      & 800,456 \\
    Elecdemand & Energy & 30T   & 1      & 17,520 \\
    Covid Mobility & Transport & D     & 362  & 148,602 \\
    Kaggle Web Traffic Weekly & Web   & W     & 145,063  & 16,537,182 \\
    Extended Web Traffic & Web   & D     & 145,063  & 370,926,091 \\
    M1 Yearly & Econ/Fin & Y     & 106     & 3,136 \\
    M1 Quarterly & Econ/Fin & Q     & 198    & 9,854 \\
    M1 Monthly & Econ/Fin & M     & 617    & 44,892 \\
    M3 Yearly & Econ/Fin & Y     & 645    & 18,319 \\
    M3 Quarterly & Econ/Fin & Q     & 756    & 37,004 \\
    M3 Monthly & Econ/Fin & M     & 1,428  & 141,858 \\
    M3 Other & Econ/Fin & Q     & 174   & 11,933 \\
    M4 Yearly & Econ/Fin & Y     & 22,739  & 840,644 \\
    M4 Quarterly & Econ/Fin & Q     & 24,000   & 2,214,108 \\
    M4 Monthly & Econ/Fin & M     & 48,000  & 10,382,411 \\
    M4 Weekly & Econ/Fin & W     & 359  & 366,912 \\
    M4 Hourly & Econ/Fin & H     & 414  & 353,500 \\
    M4 Daily & Econ/Fin & D     & 4,227 & 9,964,658 \\
    NN5 Daily & Econ/Fin & D     & 111 & 81,585 \\
    NN5 Weekly & Econ/Fin & W     & 111  & 11,655 \\
    Tourism Yearly & Econ/Fin & Y     & 419 & 11,198 \\
    Tourism Quarterly & Econ/Fin & Q     & 427  & 39,128 \\
    Tourism Monthly & Econ/Fin & M     & 366 & 100,496 \\
    CIF 2016 & Econ/Fin & M     & 72 & 6,334 \\
    Traffic Weekly & Transport & W     & 862 & 82,752 \\
    Traffic Hourly & Transport & H     & 862 & 14,978,112 \\
    Australian Electricity Demand & Energy & 30T   & 5 & 1,153,584 \\
    Rideshare & Transport & H     & 2,304 & 859,392 \\
    Saugeen & Nature & D     & 1 & 23,711 \\
    Sunspot & Nature & D     & 1 & 73,894 \\
    Temperature Rain & Nature & D     & 32,072 & 22,290,040 \\
    Vehicle Trips & Transport & D     & 329 & 32,512 \\

    \bottomrule
    \end{tabular}
  
  \label{tab:lotsa_2}%
\end{table}%
\begin{table}[ht]
  \centering
  \ContinuedFloat
  \caption{Cont. Datasets and key properties from LOTSA.}
  \begin{tabular}{lcccc}
    \toprule
    \multicolumn{5}{c}{... Continued from previous page}\\
    \midrule
    \textbf{Dataset} & \textbf{Domain} & \textbf{Frequency} & 
    \textbf{\# Time Series} & 
    \textbf{\# Obs.} \\
    \midrule
    Weather & Climate & D     & 3,010 & 42,941,700 \\
    Car Parts & Sales & M     & 2,674 & 104,286 \\
    FRED MD & Econ/Fin & M     & 107 & 76,612 \\
    Pedestrian Counts & Transport & H     & 66 & 3,130,762 \\
    Hospital & Healthcare & M     & 767 & 55,224 \\
    COVID Deaths & Healthcare & D     & 266 & 48,412 \\
    KDD Cup 2018 & Energy & H     & 270 & 2,897,004 \\
    Bitcoin & Econ/Fin & D     & 18  & 74,824 \\
    US Births & Healthcare & D     & 1 & 7,275 \\
     \midrule
    CMIP6 & Climate & 6H    & 1,351,680 & 1,973,453,000 \\
    ERA5  & Climate & H     & 245,760 & 2,146,959,000 \\
    \midrule
    Covid19 Energy & Energy & H     & 1 & 31,912 \\
    GEF12 & Energy & H     & 20 & 788,280 \\
    GEF14 & Energy & H     & 1 & 17,520 \\
    GEF17 & Energy & H     & 8 & 140,352 \\
    PDB   & Energy & H     & 1 & 17,520 \\
    Spanish & Energy & H     & 1 & 35,064 \\
    BDG-2 Hog & Energy & H     & 24 & 421,056 \\
    BDG-2 Bull & Energy & H     & 41 & 719,304 \\
    BDG-2 Cockatoo & Energy & H     & 1  & 17,544 \\
    ELF   & Energy & H     & 1 & 21,792 \\
    \midrule
    Subseasonal & Climate & D     & 862 & 14,097,148 \\
    Subseasonal Precipitation & Climate & D     & 862 & 9,760,426 \\
    \bottomrule
    \end{tabular}%
  \label{tab:lotsa_3}%
\end{table}%

\paragraph{LOTSA} Large-scale Open Time Series Archive~\citep{wooUnifiedTrainingUniversal2024} provides a large-scale collection of datasets spanning nine domains, ranging from Climate to Transport, covering a total of 231B observations. To enable the pretraining of Time Series Foundation Models (TSFMs), \citet{wooUnifiedTrainingUniversal2024} constructed and opened this extensive data archive, representing a pioneering effort in this direction. We reconstruct the timestamps using the provided start and freq (frequency) fields, and organize the data into sequences of triplets $(u,\tau,c)$ according to the variable associated with each time point. Triplets with missing values in $u$ are subsequently removed from the sequences.
\begin{table}[H]
    \centering
    \caption{Datasets and key properties from Time-IMM.}
    \begin{tabular}{llll}
    \toprule
        Dataset & Domain & \# Time Series & \# Obs. \\
        \midrule
        GDELT & Web & 8 & 193,205 \\
        RepoHealth & Health & 4 & 67,830 \\
        FNSPID & Econ/Fin & 10 & 209,688 \\
        ClusterTrace & CloudOps & 3 & 69,001 \\
        StudentLife & Human Activity  & 20 & 1,743 \\ 
        ILINet & Network & 1 & 4,918 \\ \hline
        CESNET & Network & 30 & 51,107 \\ 
        EPA-Air & Environment & 8 & 49,552 \\ 
        \bottomrule
    \end{tabular}
\end{table}
\paragraph{Time-IMM} Time-IMM~\citep{changTimeIMMDatasetBenchmark2025} provides a dataset specifically designed to capture cause-driven irregularity in multimodal multivariate time series. And it ranges from a varying taxonomy of irregularity in time series. (a) \textbf{Trigger-Based Irregularities}: GDELT and RepoHealth; (b) \textbf{Constraint-Based Irregularities}: FNSPID, ClusterTrace(different year from CloudOps LOTSA), and StudentLife; (c) \textbf{Artifact-Based Irregularities}: ILINet, CESNET, and EPA-Air.

\begin{table}[H]
    \centering
    \caption{Datasets and key properties from OWP.}
    \begin{tabular}{llll}
    \toprule
        Dataset & Domain & \# Time Series & \# Obs. \\
        \midrule
        OWP & Energy & 10 & 98,106,540 \\
        \bottomrule
    \end{tabular}
\end{table}

\paragraph{OWP} Offshore Wind Power and Wave Height measurements is a dataset released by us, consisting of 98,106,540 observations across 10 variables (Details in~\ref{tab:desc of owp}). It includes physical measurements related to offshore wind power, specifically a set of strain-gauge load components obtained through linear transformations, as well as wave height information derived from processed radar data, which captures both the magnitude and direction of ocean waves.
\begin{table}[H]
    \centering
    \caption{Description of Variables in the OWP Dataset}
    \label{tab:desc of owp}
    \small
    \begin{tabular}{llll}
    \toprule
        Variable & Unit & Description & \#Example Value \\
        \midrule
        Mtbx & kNm & Derived from tower-bottom x-axis sensor measurements & 5186.6229 \\
        Mtby & kNm & Derived from tower-bottom y-axis sensor measurements & -60491.49 \\
        Mtmx & kNm & Derived from tower-middle x-axis sensor measurements & 6149.6357 \\
        Mtmy & kNm & Derived from tower-middle x-axis sensor measurements & -33853.17 \\
        Mttx & kNm & Derived from tower-top x-axis sensor measurements & -2281.088 \\
        Mtty & kNm & Derived from tower-top y-axis sensor measurements & 18087.884 \\
        Mttz & kNm & Derived from tower-top z-axis sensor measurements & 7147.9924 \\
        Hm0\_M & cm & Wave Height & 158.24474 \\
        Th0 & deg & azimuth angle & 5.263157894736842 \\
        \bottomrule
    \end{tabular}
\end{table}

\section{Experimental Details}
\label{appendix:implementation_detail}
\subsection{UniTIMS Training Strategy}

The pre-training process is conducted on the VersaTSA archive to equip the model with universal representations under irregular temporal fluctuations, aiming to: 1) capture complex dependencies in irregular samples, and 2) enable robust zero-shot forecasting via large-scale autoregressive training. 
We implement a training strategy based on dynamic window partitioning. 
Specifically, for each training sample extracted from the long-term sequences in VersaTSA, we randomly select a partition point followed by the Beta distribution~\citep{johnson2016applicationsbetadistribution1}, splitting the patch sequence $\{p_1, p_2, \ldots, p_{m+n}\}$ into a historical prefix segment $\mathcal{P}$ and a target segment $\mathcal{T}$. 
For example, given a partition point $m$ sampled from a Beta distribution, $\mathcal{P}$ denotes the index set corresponding to the combined historical observations and covariates, i.e., $\mathcal{Y}_{t_1:t_m} \cup \mathcal{Z}_{t_1:t_{m+n}}$. The target segment $\mathcal{T}$ includes the indices of the query set $Q = \{ (t, c) \mid t \in (t_m, t_{m+n}], c \in \mathcal{C}_\mathcal{Y} \}$ and the special forecasting token \texttt{[F]}. 
During autoregressive generation, the prefix serves as the historical context where bidirectional information flow is permitted, while the target is forecasted causally, step-by-step. 
\begin{table}[htbp]
\centering

\caption{MTS and IMTS task settings with time-slicing and input configurations. $|p_m| = l_p$ denotes the same sequence length, while $\Delta T_i$ and $\Delta t_i$ represent the time durations for MTS and IMTS within each patch, respectively. The inputs for target variables $\mathcal{X}_y$ are projected alongside covariates $\mathcal{X}_z$ into a unified $d$-dimensional representation space.}

\label{tab:imts_vs_mts_chronos_style}
\resizebox{\linewidth}{!}{%
\begin{tabular}{l c c}
\toprule
\textbf{Task type} & \textbf{Time slicing} & \textbf{Inputs} \\ 
\midrule

\textbf{MTS} & 
$|p_m| = l_p$, $\Delta T_i  =\Delta T_j$
& 
$\left[ \, \underbrace{\begin{matrix} \boldsymbol{x}_1 & \dots & \boldsymbol{x}_{M_y} \\ \boldsymbol{p}_{1} & \dots & \boldsymbol{p}_{M_y} \end{matrix}}_{\text{Target Variables } \mathcal{X}_y} \ \Bigg| \ \underbrace{\begin{matrix} \boldsymbol{x}_{M_y+1} & \dots & \boldsymbol{x}_{M} \\ \boldsymbol{p}_{{M_y}+1} & \dots & \boldsymbol{p}_{M} \end{matrix}}_{\text{Covariates } \mathcal{X}_z} \, \right]\in \mathbb{R}^{2 \times M \cdot d}$ \\
\hline

\textbf{IMTS} & 
$|p_m| = l_p$, $\Delta t_i \neq \Delta t_j$
& 
$\left[ \, \underbrace{\begin{matrix} \boldsymbol{x}_1 & \dots & \boldsymbol{x}_{M_y} \\ \boldsymbol{p}_{\Delta t_1} & \dots & \boldsymbol{p}_{\Delta t_{M_y}} \end{matrix}}_{\text{Target Variables } \mathcal{X}_y} \ \Bigg| \ \underbrace{\begin{matrix} \boldsymbol{x}_{M_{y}+1} & \dots & \boldsymbol{x}_{M} \\ \boldsymbol{p}_{\Delta t_{M_{y}+1}} & \dots & \boldsymbol{p}_{\Delta t_{M}} \end{matrix}}_{\text{Covariates } \mathcal{X}_z} \, \right]\in \mathbb{R}^{2 \times M \cdot d}$ \\

\bottomrule
\end{tabular}
}
\end{table}

Crucially, our model can fit both regular and irregular MTS forecasting tasks. We summarize the task settings and input representations of MTS and IMTS under the UNITIMS framework in Table \ref{tab:imts_vs_mts_chronos_style}. For both tasks, the input sequence is divided into non-overlapping patches where each patch contains the same number of sequence points ($|p_m| = l_p$). The difference is that MTS patches span uniform time durations ($\Delta T_i = \Delta T_j$), whereas IMTS patches span varying time durations ($\Delta t_i \neq \Delta t_j$) due to non-uniform sampling. Finally, inputs from both target variables $\mathcal{X}_y$ and covariates $\mathcal{X}_z$ are projected into a unified $d$-dimensional representation space.

\subsection{Details of Evaluation Setups}
We summarize the setup for three categories of baselines here to avoid misinterpretation of the results in Sec. \ref{imts_forecasting}.
\begin{itemize}
\item IMTS-specific methods (GRU-D, mTAND, T-PatchGNN, Hi-Patch, etc.) are evaluated under the \emph{full-shot} setting, i.e., trained directly on each target dataset, and their numbers are taken from~\citep{luoHiPatchHierarchicalPatch2025} using the identical splits and metric definitions reported there.
\item TSFMs (Chronos-2, Moirai, Sundial) and {\modelname} are evaluated under the \emph{zero-shot} setting on the same splits; for IMTS inputs, regular-grid TSFMs operate on the same observed-only timestamps via their public inference interfaces without any extra fine-tuning.
\item {\modelname} is pretrained exclusively on VersaTSA, which strictly excludes the three IMTS evaluation datasets and the six regular MTS evaluation datasets (see the hold-out protocol in Sec.~\ref{sec:versatsa}).
Despite the larger pretraining scale of {\modelname} relative to full-shot baselines, all methods are evaluated on the \emph{same} test windows and use \emph{identical} MSE/MAE definitions over the observed targets, so the gains reported in Table~\ref{tab:main-result} reflect the model's representational capability rather than benefiting from extra access to evaluation data.
\end{itemize}

All experiments are implemented using PyTorch~\citep{paszke2019pytorchimperativestylehighperformance} and conducted on NVIDIA H100-NVL GPUs using TF32 precision. All models are trained with a global batch size of 1024 and with 8 attention heads. We utilize the AdamW optimizer with hyperparameters $\beta_1=0.9$, $\beta_2=0.98$, weight decay of $0.1$, and a peak learning rate of $1\times 10^{-3}$. The learning rate schedule includes a linear warmup over the first 10,000 steps, followed by cosine decay.
\textbf{Pretrain.} During the pretraining stage, we partition the dataset by time-series indices rather than constructing samples via sliding windows, which alleviates the burden on the sampler. We then randomly sample subsequences from each time series with a fixed window size of 3840, where the sampling positions are drawn from a Beta distribution with parameters $(1.6,2.4)$.
\textbf{Inference.} During inference, for any time segment whose length is insufficient to form a complete patch, we pad it accordingly and ignore the padded values in the inference process.

\subsection{Model Configurations}
\label{appendix:model_configurations}
We implement {\modelname} in three configurations: Small (60M), Base (403M), and Large (1.3B). Table~\ref{tab:model_size_detail} details the number of layers, hidden dimensions, and parameter counts of each variant.
\begin{table}[ht]
    \caption{Details of the model sizes for UniTIMS.}
    \label{tab:model_size_detail}
    \begin{center}
                    \begin{tabular}{lllll}
                        \toprule
                        ~ & Layers & $d_{model}$ & $d_{ff}$ & Params \\
                        \midrule
                        \modelof{Small} & 3 & 512 & 2048 & 60M \\
                        \modelof{Base} & 24 & 1024 & 2048 & 403M \\
                        \modelof{Large} & 48 & 1280 & 5120 & 1.3B \\
                        \bottomrule
                    \end{tabular}
    \end{center}
\end{table}

\section{Forecasting Settings}
\subsection{Evaluation Metrics}
\label{appendix:mse_and_mae}
Mean Squared Error (MSE) and MAE are commonly used to evaluate forecasting performance.
However, their computation differs slightly between irregular multivariate time series IMTS and regular multivariate time series (MTS).
In general, given a series $x$ and its prediction $\hat{x}$, the MSE and MAE are defined as:
\[
\mathrm{MSE} = \frac{1}{N}\sum_{i=1}^N(\hat{x}_i - x_i)^2, \qquad
\mathrm{MAE} = \frac{1}{N}\sum_{i=1}^N\left |\hat{x}_i - x_i \right |.
\]

Due to asynchrony and missing values, different variables (including target variables and covariates) within the same IMTS may have different numbers of observed time points. As a result, for irregularly-sampled time series (IMTS) with $C$ variates, let $\Omega_c$ denote the valid evaluated observations for variable $c$. The MSE and MAE are computed only over the observed target values and then averaged across variables:
\[
\mathrm{MSE} = \frac{1}{C}\sum_{c=1}^C \frac{1}{|\Omega_c|}\sum_{i\in\Omega_c}(\hat{x}_i^{(c)} - x_i^{(c)})^2, \qquad
\mathrm{MAE} = \frac{1}{C}\sum_{c=1}^C \frac{1}{|\Omega_c|}\sum_{i\in\Omega_c}\left |\hat{x}_i^{(c)} - x_i^{(c)} \right |.
\]
\subsection{Clarification on Autoregressive}
\label{appendix:prove_autoregressive}
In the following, we clarify how a hybrid mask can ensure autoregressive learning. Given a context sequence as the condition and a target sequence
$S = (s_1, \ldots, s_n)$, any predictive model whose conditional distribution admits the factorization can be regarded as an autoregressive forecasting model, i.e., 
\begin{equation}
P(S \mid \text{context})
= \prod_{i=1}^{n} P(s_i \mid s_{<i}, \text{context}).
\label{autoregressive}
\end{equation}

Given a context sequence as the condition and a target sequence
$S = (s_1, \ldots, s_n)$, any predictive model whose conditional
distribution admits the factorization: $
P(S \mid \text{context})
= \prod_{i=1}^{n} P(s_i \mid s_{<i}, \text{context})$
is an autoregressive forecasting model.

An inference structure is autoregressive IFF, given a sequence of length $m$, the model predicts future values by unidirectional reasoning solely based on the observed historical information, without leveraging any information from future time steps, i.e., of the following form:
\begin{equation}
    \label{eq:ar}
    P(S) = \prod_{i=1}^{m+n} P(s_i | s_{<i}) 
\end{equation}
For the inference structure formulated, its formulation is already highly consistent with the standard autoregressive form given in Eq.~\eqref{eq:ar}, except for the presence of an additional conditioning term, $context$. To further examine the role of this condition, we note that time series inputs inherently follow a temporal order. Specifically, the context sequence of length $m$ corresponds to a segment of observations that precede $s_i$ in time (here we temporarily ignore the effect of covariates). As a result, a complete history can be constructed as:
\begin{equation}
    y_{<i} = s_{<i} \cup context
\end{equation}
Under this formulation, the target $s_i$ can be regarded as a natural continuation of the historical sequence $y_{<i}$, i.e., $y_i = s_i$. Consequently, we obtain:
\begin{equation}
    P(S \mid context) = \prod_{i=1}^{m+n} P(s_i | s_{<i},context) = \prod_{i=1}^{m+n} P(y_i | y_{<i}) 
\end{equation}
\section{Supplementary Results}
\label{appendix:supplementary_results}

\subsection{Full results of regular MTS Forecasting}
\label{appendix:full_results_of_MTS}
Table \ref{tab:full_regular_zero_shot_result} evaluates the zero-shot forecasting performance of various time series foundation models on regular Multivariate Time Series (MTS) tasks. Using six benchmark datasets (ETTm1, ETTm2, ETTh1, ETTh2, ECL, and Weather) across four forecast horizons (96, 192, 336, and 720), the study utilizes MSE and MAE metrics to quantify accuracy and assess the generalization capabilities of models such as Sundial, Moirai, Chronos, and TimesFM in handling unseen data distributions and long-term dependencies.

\begin{table}[ht]
  \centering
  \caption{Full results for Regular MTS point forecasting experiments. Best results are highlighted in \textbf{bold}, and second best results are \underline{underlined}.}
  \label{tab:full_regular_zero_shot_result}
  \small
  \resizebox{\textwidth}{!}{
    \begin{tabular}{cccccccccccccccccccccc}
    \toprule
        &  & \multicolumn{2}{c}{\textbf{\modelof{Large}}} & \multicolumn{2}{c}{\textbf{\modelof{Base}}} & \multicolumn{2}{c}{\textbf{\modelof{Small}}} & \multicolumn{2}{c}{\textbf{\modelof[Sundial]{Large}}} & \multicolumn{2}{c}{\textbf{\modelof[Sundial]{Base}}} & \multicolumn{2}{c}{\textbf{\modelof[Moirai]{Large}}} & \multicolumn{2}{c}{\textbf{\modelof[Moirai]{Base}}} & \multicolumn{2}{c}{\textbf{\modelof[Chronos]{Large}}} & \multicolumn{2}{c}{\textbf{\modelof[Chronos]{Base}}} & \multicolumn{2}{c}{\textbf{TimesFM}} \\
        \midrule
        & & MSE & MAE & MSE & MAE & MSE & MAE & MSE & MAE & MSE & MAE & MSE & MAE & MSE & MAE & MSE & MAE & MSE & MAE & MSE & MAE \\ 
        \midrule
        \multirow{4}[0]{*}{ETTm1} 
        & 96  & 0.298 & \textbf{0.327} & 0.295 & 0.338 & 0.305 & 0.348 & \textbf{0.273} & \underline{0.329} & \underline{0.280} & 0.334 & 0.380 & 0.361 & 0.363 & 0.356 & 0.457 & 0.403 & 0.454 & 0.408 & 0.361 & 0.370 \\
        & 192  & 0.337 & \textbf{0.356} & 0.330 & 0.365 & 0.347 & 0.370 & \textbf{0.312} & \underline{0.357} & \underline{0.321} & 0.366 & 0.412 & 0.383 & 0.388 & 0.375 & 0.530 & 0.450 & 0.567 & 0.477 & 0.414 & 0.405 \\
        & 336  & 0.371 & \underline{0.380} & 0.375 & 0.395 & 0.405 & 0.400 & \textbf{0.343} & \textbf{0.378} & \underline{0.350} & 0.389 & 0.436 & 0.400 & 0.416 & 0.392 & 0.577 & 0.481 & 0.662 & 0.525 & 0.445 & 0.429 \\
        & 720  & 0.442 & 0.421 & 0.536 & \textbf{0.398} & 0.543 & \underline{0.406} & \underline{0.397} & 0.413 & \textbf{0.394} & 0.418 & 0.462 & 0.420 & 0.460 & 0.418 & 0.660 & 0.526 & 0.900 & 0.591 & 0.512 & 0.471 \\
        \midrule
        \multirow{4}[0]{*}{ETTm2} 
        & 96  & \textbf{0.165} & \textbf{0.253} & 0.182 & 0.262 & 0.172 & 0.259 & 0.172 & \underline{0.255} & \underline{0.170} & 0.256 & 0.211 & 0.274 & 0.205 & 0.273 & 0.197 & 0.271 & 0.199 & 0.274 & 0.202 & 0.270 \\
        & 192  & \textbf{0.217} & \textbf{0.291} & 0.235 & 0.301 & \underline{0.226} & 0.298 & 0.227 & \underline{0.296} & 0.229 & 0.300 & 0.281 & 0.318 & 0.275 & 0.316 & 0.254 & 0.314 & 0.261 & 0.322 & 0.289 & 0.321 \\
        & 336  & \textbf{0.268} & \textbf{0.328} & 0.284 & 0.335 & 0.285 & 0.334 & \underline{0.275} & \underline{0.331} & 0.281 & 0.337 & 0.341 & 0.355 & 0.329 & 0.350 & 0.313 & 0.353 & 0.326 & 0.366 & 0.360 & 0.366 \\
        & 720  & 0.362 & 0.388 & 0.383 & \textbf{0.374} & 0.449 & 0.433 & \textbf{0.343} & \underline{0.378} & \underline{0.351} & 0.387 & 0.485 & 0.428 & 0.437 & 0.411 & 0.416 & 0.415 & 0.455 & 0.439 & 0.462 & 0.430 \\
        \midrule
        \multirow{4}[0]{*}{ETTh1} 
        & 96  & 0.350 & \textbf{0.382} & 0.355 & 0.388 & 0.395 & 0.404 & \textbf{0.346} & \underline{0.383} & \underline{0.348} & 0.385 & 0.381 & 0.388 & 0.376 & 0.392 & 0.441 & 0.390 & 0.440 & 0.393 & 0.414 & 0.404 \\
        & 192  & \textbf{0.385} & \textbf{0.402} & 0.400 & 0.412 & 0.418 & 0.425 & \underline{0.386} & \underline{0.410} & 0.393 & 0.418 & 0.434 & 0.415 & 0.412 & 0.413 & 0.502 & 0.524 & 0.492 & 0.426 & 0.465 & 0.434 \\
        & 336  & \textbf{0.410} & \textbf{0.424} & 0.445 & 0.428 & 0.448 & 0.448 & \textbf{0.410} & \underline{0.426} & \underline{0.422} & 0.440 & 0.485 & 0.445 & 0.433 & 0.428 & 0.576 & 0.467 & 0.550 & 0.462 & 0.503 & 0.456 \\
        & 720  & 0.467 & 0.464 & 0.480 & 0.460 & 0.463 & 0.467 & \textbf{0.438} & \underline{0.459} & 0.481 & 0.493 & 0.611 & 0.510 & \underline{0.447} & \textbf{0.444} & 0.835 & 0.583 & 0.882 & 0.591 & 0.511 & 0.481 \\
        \midrule
        \multirow{4}[0]{*}{ETTh2} 
        & 96  & 0.285 & \textbf{0.322} & \textbf{0.268} & 0.331 & 0.300 & 0.340 & \underline{0.269} & \underline{0.330} & 0.271 & 0.333 & 0.296 & \underline{0.330} & 0.294 & \underline{0.330} & 0.320 & 0.345 & 0.308 & 0.343 & 0.315 & 0.349 \\
        & 192  & 0.335 & \underline{0.365} & \textbf{0.324} & 0.372 & 0.330 & \textbf{0.360} & \underline{0.325} & 0.373 & 0.327 & 0.376 & 0.361 & 0.371 & 0.365 & 0.375 & 0.406 & 0.399 & 0.384 & 0.392 & 0.388 & 0.395 \\
        & 336  & 0.365 & 0.392 & \underline{0.356} & \textbf{0.389} & 0.370 & \underline{0.390} & \textbf{0.354} & 0.400 & \textbf{0.354} & 0.402 & 0.390 & \underline{0.390} & 0.376 & \underline{0.390} & 0.492 & 0.453 & 0.429 & 0.430 & 0.422 & 0.427 \\
        & 720  & 0.423 & 0.425 & 0.456 & \textbf{0.404} & 0.416 & 0.422 & \underline{0.389} & 0.443 & \textbf{0.381} & 0.435 & 0.423 & \underline{0.418} & 0.416 & 0.433 & 0.603 & 0.511 & 0.501 & 0.477 & 0.443 & 0.454 \\
        \midrule
        \multirow{4}[0]{*}{ECL} 
        & 96  & \textbf{0.127} & \textbf{0.223} & 0.131 & \underline{0.226} & 0.136 & 0.235 & \underline{0.130} & 0.227 & 0.132 & 0.229 & 0.153 & 0.241 & 0.160 & 0.250 & 0.152 & 0.229 & 0.154 & 0.231 & - & - \\
        & 192  & \textbf{0.144} & \textbf{0.241} & \underline{0.149} & \underline{0.244} & 0.156 & 0.255 & 0.150 & 0.247 & 0.152 & 0.250 & 0.169 & 0.255 & 0.175 & 0.263 & 0.172 & 0.250 & 0.179 & 0.254 & - & - \\
        & 336  & \textbf{0.164} & \textbf{0.261} & \underline{0.169} & \underline{0.264} & 0.176 & 0.275 & 0.170 & 0.268 & 0.173 & 0.271 & 0.187 & 0.273 & 0.187 & 0.277 & 0.203 & 0.276 & 0.214 & 0.284 & - & - \\
        & 720  & \underline{0.213} & \textbf{0.302} & 0.219 & \underline{0.306} & \textbf{0.208} & 0.311 & 0.214 & 0.307 & 0.218 & 0.311 & 0.237 & 0.313 & 0.228 & 0.309 & 0.289 & 0.337 & 0.311 & 0.346 & - & - \\
        \midrule
        \multirow{4}[0]{*}{Weather} 
        & 96  & 0.162 & \underline{0.208} & 0.170 & 0.215 & \underline{0.160} & 0.210 & \textbf{0.157} & \underline{0.208} & \textbf{0.157} & \textbf{0.205} & 0.199 & 0.211 & 0.220 & 0.217 & 0.194 & 0.235 & 0.203 & 0.238 & - & - \\
        & 192  & 0.211 & \underline{0.255} & 0.225 & \underline{0.255} & 0.212 & \underline{0.255} & \underline{0.207} & 0.256 & \textbf{0.205} & \textbf{0.251} & 0.246 & \textbf{0.251} & 0.271 & 0.259 & 0.249 & 0.285 & 0.256 & 0.290 & - & - \\
        & 336  & 0.262 & \underline{0.293} & 0.270 & 0.295 & 0.270 & 0.295 & \underline{0.259} & 0.295 & \textbf{0.253} & \textbf{0.289} & 0.274 & 0.291 & 0.286 & 0.297 & 0.302 & 0.327 & 0.314 & 0.336 & - & - \\
        & 720  & \underline{0.329} & \underline{0.340} & 0.351 & 0.351 & 0.402 & 0.372 & 0.327 & 0.342 & \textbf{0.320} & \textbf{0.336} & 0.337 & \underline{0.340} & 0.373 & 0.354 & 0.372 & 0.378 & 0.397 & 0.396 & - & - \\
        \midrule
        \multirow{2}{*}{{Count}} & {1-st} & 8 & 13 & 2 & 4 & 1 & 1 & 9 & 1 & 7 & 4 & 0 & 1 & 0 & 0 & 0 & 0 & 0 & 0 & 0 & 0 \\
        & {2-nd} & 2 & 6 & 3 & 5 & 2 & 3 & 9 & 12 & 7 & 0 & 0 & 4 & 1 & 2 & 0 & 0 & 0 & 0 & 0 & 0 \\
        \bottomrule
    \end{tabular}}
\end{table}

\newpage

\subsection{Varying Context and Horizon of IMTS Forecasting}
\label{appendix:varying_context_horizon}
On the Human Activity dataset (Table~\ref{tab:ha_varying_horizon}), we evaluate model performance under varying observation-to-forecast ratios, comparing general Transformers against specialized architectures. Across both the balanced (2000ms $\to$ 2000ms) and the challenging short-history (1000ms $\to$ 3000ms) configurations, the results quantify the superior adaptability of our proposed variants to shifting temporal windows.
\begin{table}[t]
    \centering
    \caption{Performance of varying observation and forecast horizons on Human Activity dataset. Best results are highlighted in \textbf{bold}, and second best results are \underline{underlined}.}
    \label{tab:ha_varying_horizon}
    \begin{tabular}{lcccc}
        \toprule
          ~ & \multicolumn{2}{c}{2000ms $\rightarrow$ 2000ms} & \multicolumn{2}{c}{1000ms $\rightarrow$ 3000ms} \\
        \cmidrule(lr){2-3} \cmidrule(lr){4-5}
        ~ & MSE $(\times 10^{-3})$ & MAE $(\times 10^{-2})$ & MSE $(\times 10^{-3})$ & MAE $(\times 10^{-2})$ \\
        \midrule
        \textbf{PatchTST} & 7.25$\pm$0.29 & 6.26$\pm$0.17 & 8.97$\pm$1.96 & 6.94$\pm$0.94 \\
        \textbf{iTransformer} & 7.49$\pm$4.72 & 6.08$\pm$2.17 & 5.58$\pm$0.04 & 5.13$\pm$0.05 \\
        \textbf{TimesNet} & 5.38$\pm$0.30 & 5.32$\pm$0.20 & 9.90$\pm$0.42 & 7.34$\pm$1.81 \\
        \textbf{TimeMixer} & 5.39$\pm$0.54 & 5.05$\pm$0.38 & 5.96$\pm$0.19 & 5.36$\pm$0.10 \\
        \midrule
       \textbf{GRU-D} & 5.93$\pm$0.10 & 5.66$\pm$0.66 & 6.14$\pm$0.76 & 5.75$\pm$0.49 \\
        \textbf{CRU} & 4.12$\pm$0.08 & 4.43$\pm$0.06 & 4.85$\pm$0.09 & 4.86$\pm$0.07 \\
        \textbf{mTAND} & 4.38$\pm$0.37 & 4.59$\pm$0.29 & 5.29$\pm$0.32 & 5.12$\pm$0.23 \\
        \textbf{NeuralFlow} & 5.47$\pm$0.49 & 5.35$\pm$0.28 & 6.01$\pm$0.91 & 5.66$\pm$0.60 \\
        \textbf{Latent-ODE} & 5.04$\pm$0.46 & 5.11$\pm$0.29 & 5.48$\pm$0.21 & 5.33$\pm$0.14 \\
        \textbf{tPatchGNN} & 3.71$\pm$0.20 & 3.89$\pm$0.16 & 4.56$\pm$0.08 & 4.32$\pm$0.06 \\
        \textbf{Hi-Patch} & 3.29$\pm$0.04 & 3.70$\pm$0.04 & 4.21$\pm$0.08 & 4.25$\pm$0.07 \\
        \midrule
        \textbf{{\modelof{Small}}} & 3.08$\pm$0.03 & 3.53$\pm$0.04 & 3.97$\pm$0.06 & 4.04$\pm$0.07 \\
        \textbf{{\modelof{Base}}} & \textbf{2.14$\pm$0.02} & \textbf{1.89$\pm$0.03} & \textbf{2.83$\pm$0.02} & \textbf{2.32$\pm$0.03} \\
        \textbf{{\modelof{Large}}} & \underline{2.84$\pm$0.03} & \underline{2.15$\pm$0.06} & \underline{3.74$\pm$0.03} & \underline{2.70$\pm$0.05} \\
        \bottomrule
    \end{tabular}
\end{table}

Similarly, on the PhysioNet dataset (Table~\ref{tab:physionet_varying_horizon}), the Base model achieves state-of-the-art results, significantly outperforming both general Transformers and specialized baselines like Hi-Patch. It demonstrates remarkable robustness when transitioning to the long-prediction (12h $\to$ 36h) scenario, effectively capturing irregularly sampled physiological features. Collectively, these findings validate the model's strong generalization capabilities and its efficacy in handling data-constrained environments across diverse domains.
\begin{table}[t]
    \centering
    \caption{Performance of varying observation and forecast horizons on PhysioNet dataset. Best results are highlighted in \textbf{bold}, and second best results are \underline{underlined}.}
    \label{tab:physionet_varying_horizon}
    \begin{tabular}{lcccc}
        \toprule
          ~ & \multicolumn{2}{c}{36h $\rightarrow$ 12h} & \multicolumn{2}{c}{12h $\rightarrow$ 36h} \\
        \cmidrule(lr){2-3} \cmidrule(lr){4-5}
       ~ & MSE $(\times 10^{-3})$ & MAE $(\times 10^{-2})$ & MSE $(\times 10^{-3})$ & MAE $(\times 10^{-2})$ \\
        \midrule
        \textbf{PatchTST} & 26.13$\pm$1.75 & 11.25$\pm$0.29 & 25.63$\pm$1.51 & 10.7$\pm$0.38 \\ 
        \textbf{iTransformer} & 56.83$\pm$21.17 & 17.05$\pm$5.21 & 54.17$\pm$17.91 & 17.03$\pm$4.18 \\ 
        \textbf{TimesNet} & 9.43$\pm$0.53 & 5.55$\pm$0.20 & 9.26$\pm$0.18 & 5.58$\pm$0.10 \\ 
        \textbf{TimeMixer} & 12.52$\pm$0.45 & 6.47$\pm$0.21 & 19.86$\pm$0.17 & 8.38$\pm$0.18 \\ 
        \midrule
        \textbf{GRU-D} & 6.85$\pm$0.37 & 4.88$\pm$0.18 & 7.80$\pm$0.22 & 5.13$\pm$0.13 \\ 
        \textbf{CRU }& 6.74$\pm$0.21 & 4.82$\pm$0.11 & 7.66$\pm$0.14 & 4.97$\pm$0.05 \\ 
        \textbf{mTAND} & 5.61$\pm$0.31 & 4.15$\pm$0.09 & 7.46$\pm$0.19 & 4.85$\pm$0.05 \\ 
        \textbf{NeuralFlow} & 8.87$\pm$1.00 & 5.43$\pm$0.18 & 7.98$\pm$0.57 & 5.08$\pm$0.24 \\ 
        \textbf{Latent-ODE} & 6.99$\pm$0.24 & 4.74$\pm$0.11 & 7.28$\pm$0.13 & 4.83$\pm$0.07 \\ 
        \textbf{tPatchGNN} & 4.22$\pm$0.09 & 3.38$\pm$0.04 & 6.45$\pm$0.11 & 4.24$\pm$0.09 \\ 
        \textbf{Hi-Patch} & 4.16$\pm$0.08 & 3.31$\pm$0.06 & 6.30$\pm$0.06 & 4.12$\pm$0.05 \\ 
        \midrule
        \textbf{{\modelof{Small}}} & 4.52$\pm$0.08 & 3.7$\pm$0.08 & 6.41$\pm$0.07 & 4.38$\pm$0.05 \\ 
        \textbf{{\modelof{Base}}} & \textbf{3.93$\pm$0.04} & \textbf{2.41$\pm$0.03} & \textbf{6.06$\pm$0.04} & \textbf{4.04$\pm$0.03} \\ 
        \textbf{{\modelof{Large}}} & \underline{4.11$\pm$0.03} & \underline{2.72$\pm$0.05} & \underline{6.21$\pm$0.06} & \underline{4.03$\pm$0.05} \\ 
        \bottomrule
    \end{tabular}
\end{table}

\subsection{Benchmark Results on fev-bench}
\label{appendix:fev_bench}
We evaluate our model on the open leaderboard
fev-bench, which consists of 100 forecasting tasks and offers the most comprehensive coverage of diverse
real-world scenarios, including tasks with covariates. None of these datasets or tasks were seen during the training phase.
Figure~\ref{fig:fev-bench} shows the results on fev-bench with respect to the Agg. Relative WQL and Agg. Relative MASE metrics, which evaluate the point forecast accuracy (the precision of the model's mean or median (point) predictions under a scale-invariant lens) and the generalization over baselines (whether the foundation model has actually learned complex temporal patterns, like complex seasonality and trends, that give it a true edge over trivial, statistical baselines). From the results, we observe that all pre-trained models under zero-shot inference show significant superiority in scalability compared with task-specific models under supervised training and statistical models. Meanwhile, TabPFN-TS stood out because its native table-form feature extraction design can easily align with the covariance-rich characteristics of fev-bench. Despite being designed for irregular time series, our model still demonstrates robust and trustworthy performance on this benchmark.

\subsection{Inference Time}
\label{appendix:inference_time}
\begin{figure}[t]
    \centering
    \includegraphics[width=\columnwidth]{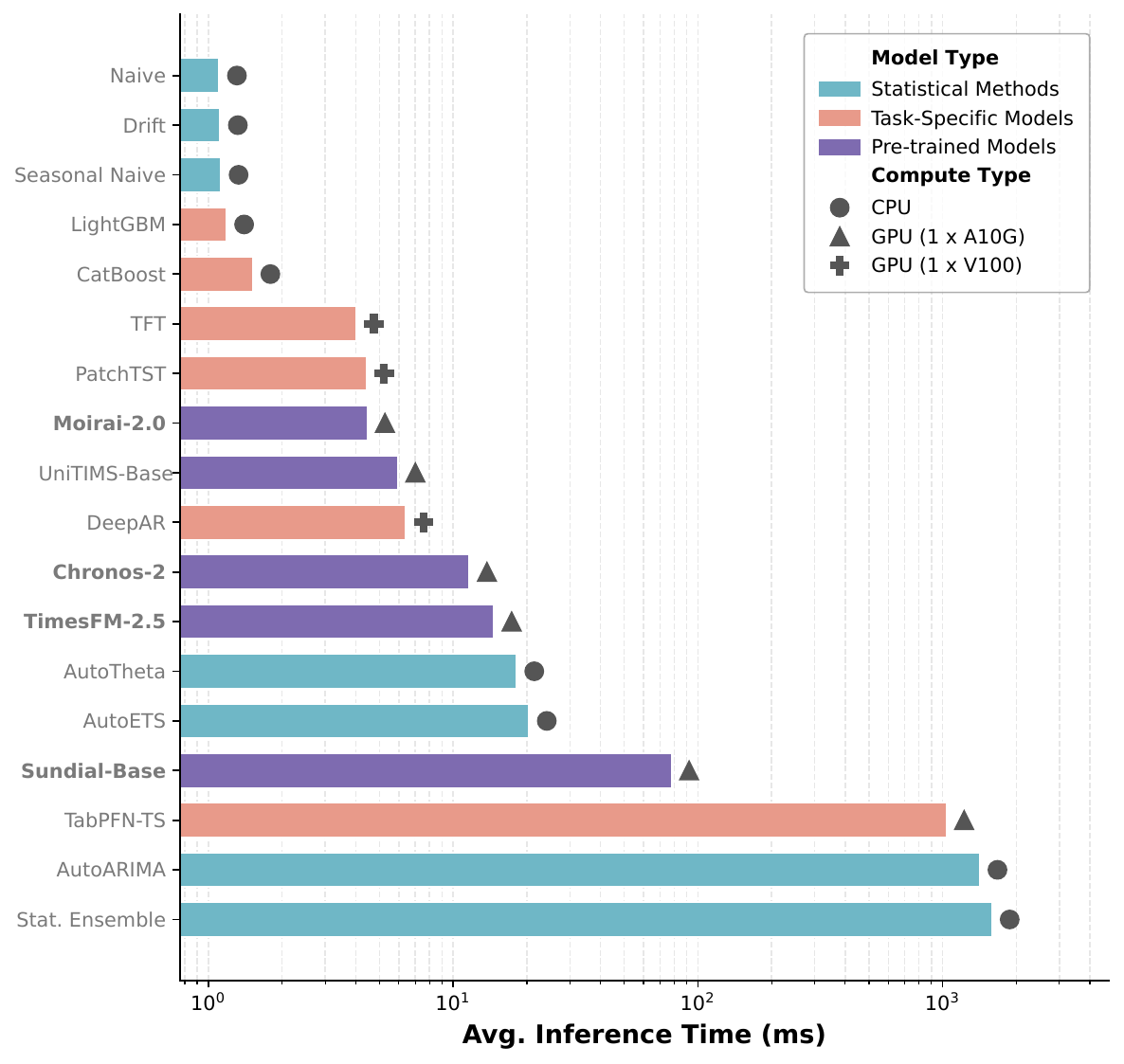}
    \caption{Inference time comparison across representative TSFMs. All models are evaluated under the same hardware setup.}
    \label{fig:inference_time}
\end{figure}

To further evaluate the practical deployment efficiency of {\modelname}, we compare the inference time of representative TSFMs under the same hardware configuration. Figure~\ref{fig:inference_time} reports the wall-clock inference time across different models. {\modelname} achieves competitive inference speed relative to other foundation models of comparable scale, demonstrating that our time-aware hybrid attention design does not introduce prohibitive computational overhead despite modeling irregular temporal structures explicitly.

\subsection{Training Dynamics}
\label{appendix:training_dynamics}
\begin{figure}[ht]
    \centering
    \begin{minipage}[t]{0.49\linewidth}
        \vspace{0pt}
        \centering
        \includegraphics[width=\linewidth]{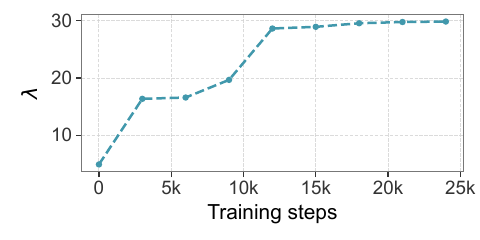}
        \caption{Evolution of $\lambda$ during training.}
        \label{fig:lambda_value}
    \end{minipage}\hfill
    \begin{minipage}[t]{0.49\linewidth}
        \vspace{0pt}
        \centering
        \includegraphics[width=\linewidth]{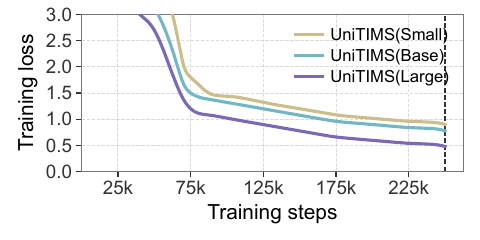}
        \caption{Training loss across model sizes.}
        \label{fig:training_loss}
    \end{minipage}
\end{figure}

The $\lambda$ value in Time Attention Bias reflects the extent to which temporal similarity modifies the model's attention. A higher $\lambda$ value indicates a greater degree of adjustment to attention scores by temporal similarity. Figure~\ref{fig:lambda_value} illustrates the variation in $\lambda$ with increasing training steps during the pre-training process of the \modelof{Small} model, with sampling conducted every 3,000 steps. It can be observed that as the model's training steps increase, the degree of correction applied by temporal similarity to the model grows, gradually trending towards a stable state in the later stages of training. The curve demonstrates that an optimal weight can be found as a bound of using temporal information in the universal representation of IMTS when the model converges.

We further investigate how the model's effectiveness evolves when the number of parameters scales up. Figure~\ref{fig:training_loss} presents the training curves for \modelname of different sizes. Compared to \modelname(Small), the large variant converges to a 24.4\% lower training loss, indicating that increased model capacity leads to more effective optimization and stronger representational capability in generative forecasting.
We also note that on PhysioNet, \modelof{Base} marginally outperforms \modelof{Large} in MSE/MAE (Table~\ref{tab:main-result}), which appears to invert the scaling trend observed in pretraining loss. We attribute this to the small evaluation size and high label sparsity of PhysioNet: under the same downstream evaluation setting, the larger variant exhibits higher generalization variance on small held-out IMTS targets, while still attaining the lowest pretraining loss. Across the larger Human Activity and USHCN benchmarks, \modelof{Large} consistently delivers the best results, confirming that the scaling benefit holds in the regime where evaluation noise is comparatively small. We regard mitigating this small-dataset variance (e.g., via downstream-aware early stopping or temperature-controlled decoding) as a practical limitation worth further study.

\section{Forecast Visualizations}
\label{appendix:forecast_visualizations}
\begin{figure}[H]
    \centering
    \includegraphics[width=\linewidth]{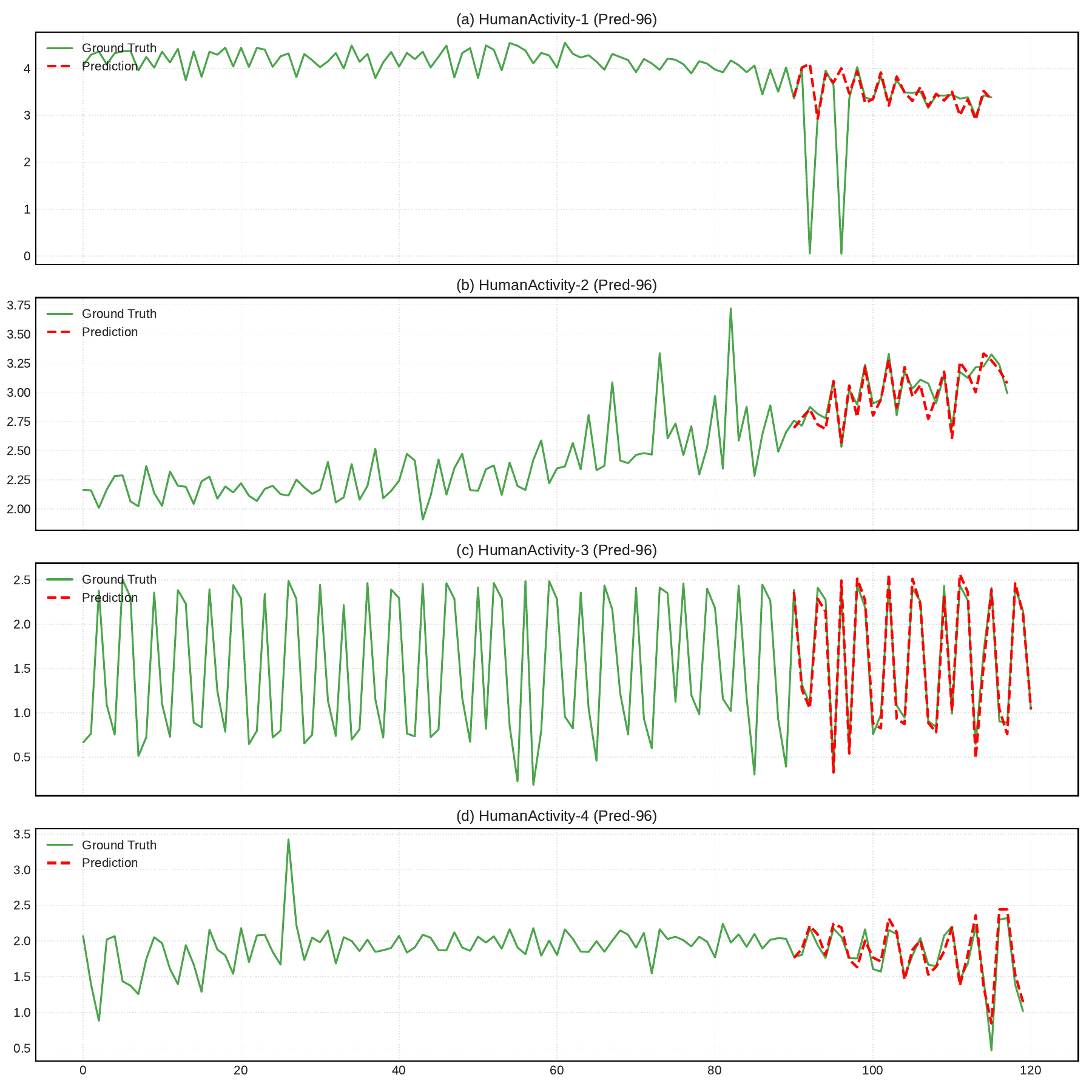}
    \caption{Visualizations of zero-shot forecasts from {\modelof{Large}} on the start segments of the Human Activity dataset.}
    \label{fig: IMTS_vis}
\end{figure}

\begin{figure}[H]
    \centering
    \includegraphics[width=\linewidth]{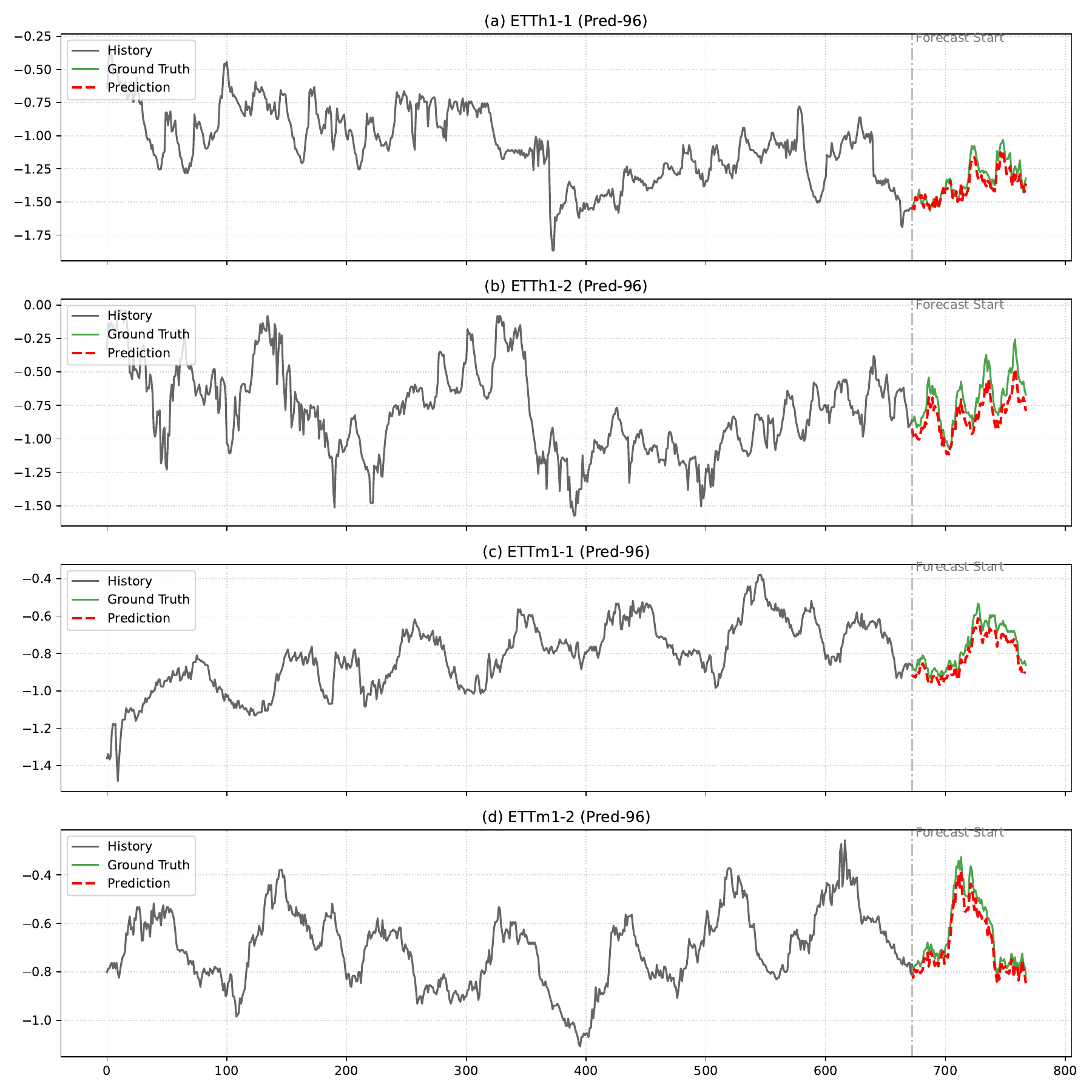}
    \caption{Visualizations of zero-shot forecasts from {\modelof{Large}} on regular MTS datasets (ETTh1 and ETTm1).}
    \label{fig: ETT_vis}
\end{figure}

Figure~\ref{fig: IMTS_vis} and Figure~\ref{fig: ETT_vis} present qualitative zero-shot forecasting results of UniTIMS-Large on IMTS and regular MTS datasets, respectively. In Figure~\ref{fig: IMTS_vis}, four representative cases from the Human Activity dataset exhibit diverse temporal behaviors, including stable variations, gradual trends, and periodic fluctuations. Under irregular sampling, the predicted curves generally preserve the overall evolution of the ground truth and capture the major temporal patterns. Although abrupt outliers or sharp local changes are not fully recovered, the model maintains stable predictions without obvious trend deviation. Figure~\ref{fig: ETT_vis} further evaluates the transferability of UniTIMS-Large on regular MTS datasets, including ETTh1 and ETTm1. The predictions are smoothly connected with the historical context and remain well aligned with the future ground truth, covering both slowly varying trends and relatively high-frequency local fluctuations.
Overall, these visualizations provide intuitive evidence that UniTIMS-Large achieves robust zero-shot forecasting across heterogeneous time-series scenarios, including irregular intervals, asynchronous observations, varying frequencies, trends, and periodic patterns.
\clearpage
\subsection{IMTS}

\clearpage
\subsection{MTS}
\clearpage

\end{document}